\documentclass[letterpaper, 10 pt, conference]{ieeeconf}  

\usepackage[noadjust]{cite}

\usepackage{xcolor}
\usepackage{graphicx}
\usepackage{amsmath}
\usepackage{multirow}
\usepackage{booktabs}
\usepackage{amssymb}
\usepackage{float}
\usepackage{url}
\usepackage{colortbl} 
\let\labelindent\relax
\usepackage[shortlabels]{enumitem}
\usepackage{hyperref}
\usepackage[font=small]{caption}

\newcommand{\modelname}[0]{DreamControl}

\newcommand{\parahead}[1]{\noindent\textbf{#1}:\ }

\IEEEoverridecommandlockouts                              

\title{\LARGE \bf
\modelname: Human-Inspired Whole-Body Humanoid Control for Scene Interaction via Guided Diffusion
}



\author{
    Dvij Kalaria\textsuperscript{1,2} \and Sudarshan Harithas\textsuperscript{1,3} \and Pushkal Katara\textsuperscript{1} \and Sangkyung Kwak\textsuperscript{1} \and
    Sarthak Bhagat\textsuperscript{1} \and S. Shankar Sastry\textsuperscript{2} \and \qquad Srinath Sridhar\textsuperscript{3} \and Sai Vemprala\textsuperscript{1} \and
    Ashish Kapoor\textsuperscript{1} \and Jonathan Huang\textsuperscript{1}
\thanks{$^{1}$General Robotics; This work was performed while Dvij Kalaria and Sudarshan Harithas were at General Robotics.}%
\thanks{$^{2}$University of California, Berkeley}
\thanks{$^{3}$Brown University}
}

\begin{document}

\maketitle

\thispagestyle{empty}
\pagestyle{empty}

\begin{abstract}

We introduce \modelname, a novel methodology for learning autonomous whole-body humanoid skills. \modelname~leverages the strengths of diffusion models and Reinforcement Learning (RL): our core innovation is the use of a diffusion prior trained on human motion data, which subsequently guides an RL policy in simulation to complete specific tasks of interest (e.g., opening a drawer or picking up an object). We demonstrate that this human motion-informed prior allows RL to discover solutions unattainable by direct RL, and that diffusion models inherently promote natural-looking motions, aiding in sim-to-real transfer. We validate \modelname's effectiveness on a Unitree G1 robot across a diverse set of challenging tasks involving simultaneous lower and upper body control and object interaction.

\end{abstract}

\vspace{1mm}
\hspace{-4mm} \textbf{Project website:} \href{https://genrobo.github.io/DreamControl/}{https://genrobo.github.io/DreamControl/}

\section{Introduction}


Significant advancements in humanoid robot control have been made in recent years, particularly in locomotion and motion tracking, leading to impressive demonstrations such as robot dancing~\cite{cheng2024express,ji2024exbody2} and kung-fu~\cite{xie2025kungfubot}. However, for humanoid robots to transition from mere exhibitions to universal assistants, they must be able to interact with their environment by fully leveraging their humanoid form factor's mobility and extensive range of motion. This includes tasks such as stooping to pick up objects, squatting for heavy boxes, bracing to open drawers or doors, and precise pushing, punching, or kicking of specific targets.

These tasks are sometimes referred to as whole-body manipulation and loco-manipulation tasks, and continue to pose substantial challenges for the humanoid robotics field. Existing approaches to humanoid manipulation often simplify the problem by fixing the lower body (e.g.,~\cite{lin2025sim}), training upper and lower bodies separately with the lower body reacting to the upper~(e.g.,~\cite{li2025amo}), or focusing exclusively on computer graphics applications~(e.g.,~\cite{pan2025tokenhsi,tevet2025closd}).

A major challenge in whole-body loco-manipulation is that of contending with multiple timescales. First, there is the problem of dynamically maintaining stability and balance,
which requires short-horizon control and robustness at the sub-second scale and is challenging due to high degrees of freedom, underactuation, and a high center of mass.
Recent approaches address this part of the problem
with reinforcement learning (RL) and sim-to-real transfer. 

Concurrently, the robot needs to formulate a motion plan for grasping distant objects, which is a long-horizon problem, spanning up to tens of seconds. The long-horizon and high-dimensional nature of bimanual manipulation leads to a particularly challenging RL exploration problem, requiring complex and precise coordination between both sets of arms and hands. Directly applying RL in such scenarios can therefore often fail or lead to unnatural behaviors that generalize poorly to the real world~\cite{luo2024omnigrasp}. 

\begin{figure}[t!]
    \centering
    \includegraphics[
        trim={0 0 5.2in 0}, 
        clip,
        width=0.9\columnwidth 
    ]{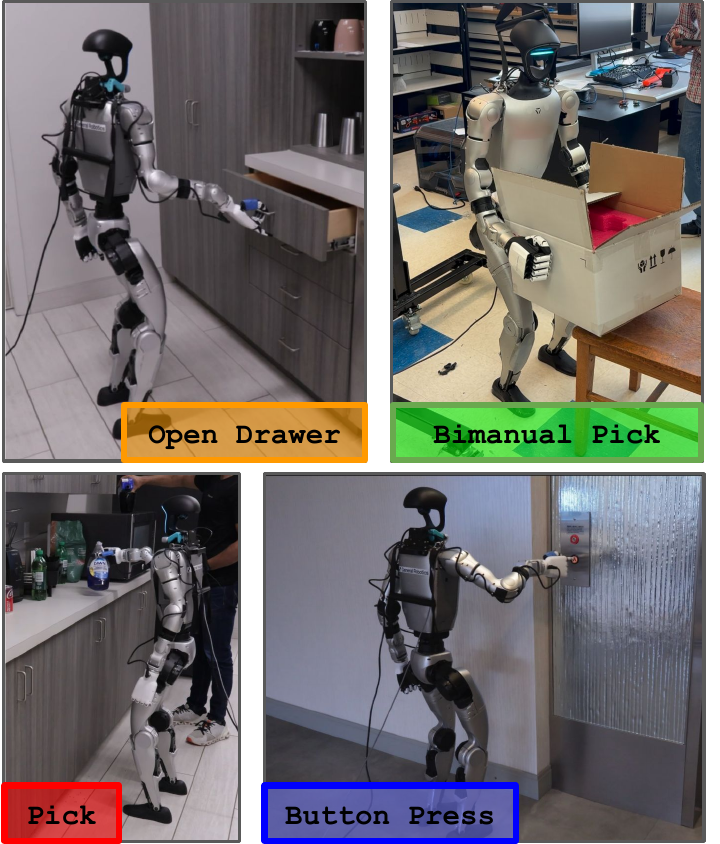}
    \caption{Unitree G1 humanoid performing various skills trained via ~\modelname,
    including (1) opening a drawer, (2) bimanual pick (of a box), (3) ordinary pick and (4) pressing an elevator button.}\vspace{-3mm}
    \label{fig:splash}
    \vspace{-0.1in}
\end{figure}

Consequently, modern approaches often rely on real-world data collection and imitation learning.  Among these approaches, diffusion policies~\cite{chi2023diffusion} (and related flow matching based approaches~\cite{black2410pi0}) have shown promise in generating long, consistent temporal data, offering a potential solution to these challenges. 
Conceptually, diffusion-based approaches are a natural fit for the multimodal nature of 
action distributions in manipulation and also scale well, allowing for learning multiple tasks simultaneously.
A complication, however, is the limited availability of  teleoperation data for whole-body humanoid control, leading some groups to propose using only upper-body teleoperation data. Whatever the form, however, collecting large teleoperation data can be labor-intensive and difficult to scale (Goldberg~\cite{goldberg2025good} refers to this as the 100,000-year data gap in robotics).

We introduce \emph{\modelname}, a two-stage methodology for learning autonomous whole-body skills that explicitly addresses the above issues by leveraging the strengths of both diffusion models and RL. Our key innovation is the use of a diffusion prior over human motions, specifically utilizing OmniControl~\cite{xie2024omnicontrol}, which takes text conditions (e.g., ``open the drawer'') and spatiotemporal guidance (e.g., enforcing a wrist position at a specific time) as input. 
Subsequently, we retarget motion samples from this prior to the robot form factor of interest and train an RL policy in simulation to follow these retargeted samples while simultaneously completing some task of interest (e.g., lifting a heavy box). We demonstrate that both privileged and non-privileged versions of this policy can be trained with minor modifications, facilitating convenient deployment to real robots.

Our approach offers several benefits. First, instead of relying on teleoperation data, it only depends on human data for training the diffusion prior. Human motion data is far more abundant (e.g. from motion capture and video sources), and since it only informs the prior, we do not depend on access to explicit reference trajectories during policy rollouts, enabling fully autonomous task execution. We show that this prior enables RL to discover solutions unattainable by direct RL approaches. 
Additionally, our diffusion prior contributes to bridging the sim-to-real gap by proposing natural-looking (less robotic) motion plans that generally do not include extreme motions. 

We demonstrate the success of \modelname~on a Unitree G1 robot across a variety of challenge
tasks, including those emphasizing simultaneous lower and upper body control and object interaction, alongside ablations that validate our design choices.


\section{Related work}
Our work is inspired by three main strands of research: robot manipulation (with imitation learning as well as on-policy RL), RL for legged robots (from locomotion and teleoperation to full autonomy) as well as the character animation and human motion modeling literatures.

\subsection{Recent Advances in Manipulation}
Modern deep learning approaches to robot manipulation are commonly based on imitation learning~\cite{vuong2023open,o2024open,kim2024openvla}. Our work draws particularly on those that leverage diffusion~\cite{sohl2015deep,watson2021learning} or related flow matching~\cite{lipman2022flow} approaches to policy parameterization~\cite{chi2023diffusion,team2024octo,black2410pi0,bjorck2025gr00t,barreiros2025careful,wen2025dexvla,liu2024rdt}.  These approaches attempt to emulate the success of LLMs because they scale well given lots of data, but unlike text, robot data is not ubiquitously available on the internet.  Collecting robot trajectories is costly, requiring expensive teleoperation rigs as well as training and paying human teleoperators. 

There are also on-policy RL approaches trained in simulated environments that are more scalable~\cite{li2025maniptrans,lin2025sim} — though robust sim2real transfer is challenging. Most relevant to our approach is the work of  Lin et al~\cite{lin2025sim} who demonstrate robust bimanual manipulation skills on a humanoid robot, but do not address whole body skills. Like~\cite{lin2025sim} we use 
on-policy RL (instead of behavior cloning with teleoperated trajectories) --- but our models
are informed by a diffusion prior over human motion,  significantly reducing
the need for reward engineering.

\subsection{RL controllers for legged robots}
In recent years, deep RL has seen significantly increased adoption in RL controllers for legged robots, starting with robust legged locomotion policies for quadrupeds~\cite{kumar2021rma,zhuang2023robot,cheng2023parkour} followed by bipedal form factors (including humanoids)~\cite{kumar2022adapting,zhuang2024humanoid,radosavovic2024real,van2024revisiting,seo2025fasttd3,he2025attention,videomimic}. More recently authors have proposed whole body motion tracking and teleoperation approaches which allow a robot to track the motion of a human teleoperator~\cite{cheng2024express,ji2024exbody2,he2024learning,he2024omnih2o,ze2025twist,he2024hover,dugar2024mhc,chen2025gmt,li2025clone,zhang2025falcon} including advances in handling agile and extreme motions (e.g. KungFuBot~\cite{xie2025kungfubot} and ASAP~\cite{he2025asap}).  See also ~\cite{gu2025humanoid} for a more complete overview of the field.

Finally, beyond tracking a provided human motion, lies the challenge of enabling fully autonomous execution of specific tasks, e.g. kicking, sitting, swinging a golf club (we will sometimes refer to these as ``skills'')~\cite{zhang2025hub,mao2024learning,he2025asap,xue2025leverb,fu2024humanplus,li2025amo,liao2025beyondmimic,zhang2025unleashinghumanoidreachingpotential}.

Among these works, HumanPlus~\cite{fu2024humanplus} and AMO~\cite{li2025amo} demonstrate whole body autonomous
task execution but require teleoperated trajectories for IL.  R2S2~\cite{zhang2025unleashinghumanoidreachingpotential} train a limited
set of ``primitive'' skills and focus primarily on ensembling these primitives using
IL and RL --- whereas our focus is on a recipe for training a library of such primitive skills.
Finally we note that  BeyondMimic~\cite{liao2025beyondmimic} also leverages both
guided diffusion and RL, but the way that diffusion is used is mostly 
orthogonal to our work.  Guidance in their diffusion policy is ``coarse'' rather than fine-grained compared to our work and does not account for object interaction or long range planning.

\subsection{Character Animation and Motion Models}
There is also a similar literature on modeling the movement of humanoids in physically realistic character animation settings~\cite{peng2018deepmimic,peng2021amp,peng2022ase,luo2023perpetual,luo2023universal,luo2024omnigrasp,wang2025skillmimic,luo2024smplolympics,tirinzoni2025zero}.  
By having access to privileged simulation states and no sim-to-real distribution shifts,
solving problems in this simplified synthetic setting first has proven useful as a stepping stone prior to crossing the sim-to-real gap.

We are in particular influenced by statistical priors over human motion --- which have a rich history (see e.g. ~\cite{brand2000style,li2002motion,wang2007gaussian}) and today leverage the recent advances in generative AI (such as diffusion models and autogregressive transformers) ~\cite{tevet2023human,tevet2025closd,xie2024omnicontrol,xu2025intermimic,pan2025tokenhsi}.

\begin{figure*}[t!]
    \centering
    \includegraphics[
        width=1.\textwidth 
    ]{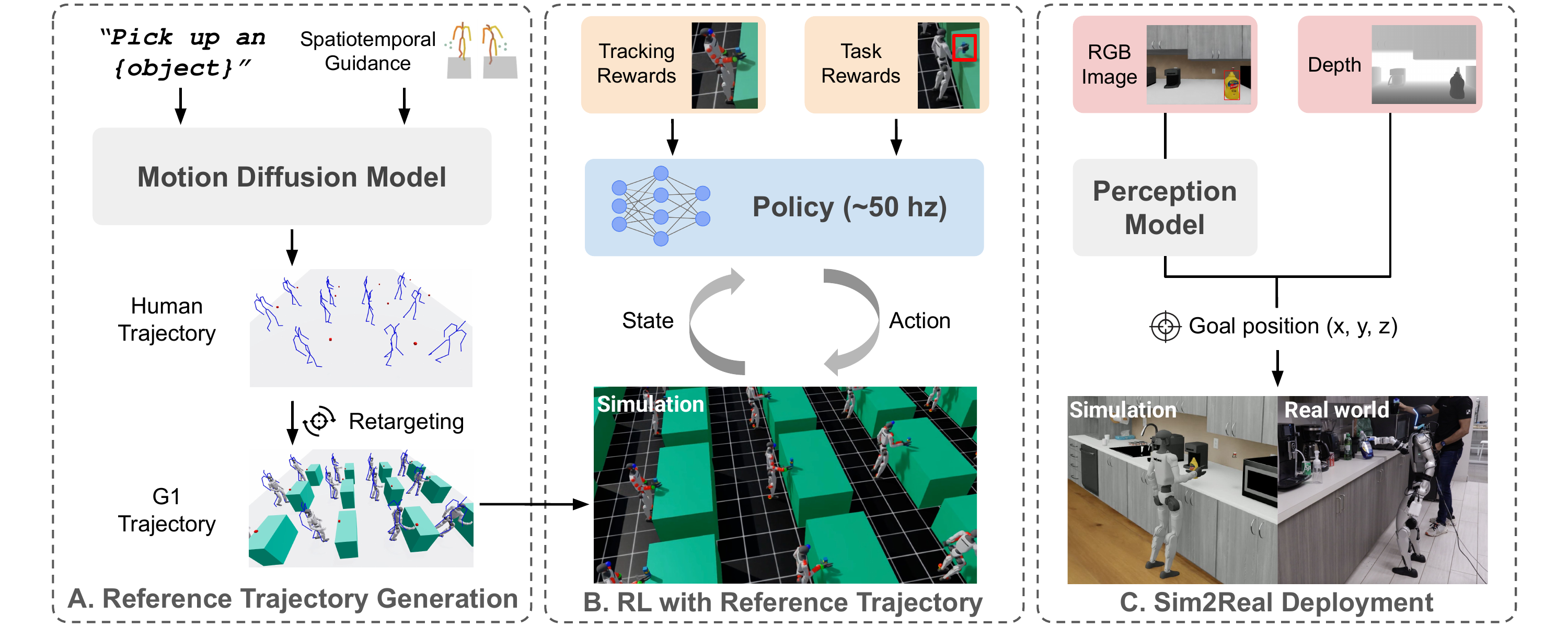}
    \vspace{-0.12in}
    \caption{\textbf{\modelname\,Overview}: 
    (A) we first generate text and spatiotemporally guided human motion trajectories using diffusion; 
    (B) we train goal-conditioned RL policies to track these generated trajectories while completing some task of interest; 
    (C) we deploy these policies to a real humanoid, leveraging off-the-shelf vision models to determine spatial guidance inputs for the RL policy.
    }
    \label{fig:main}
    \vspace{-0.1in}
\end{figure*}

Among these papers, our work is most influenced by OmniGrasp~\cite{luo2024omnigrasp}, CloSd~\cite{tevet2025closd} and TokensHSI~\cite{pan2025tokenhsi}, all of which explicitly handle object/scene interactions.  Omnigrasp  leverages a prior over human motions (PULSE,~\cite{luo2023universal}) taking the form of a bottleneck VAE that directly predicts actions though has the disadvantage of being
somewhat more awkward to interpret directly as a prior on human trajectories.  
CloSd generates motion plans via diffusion and using an RL-trained policy to 
execute in simulation.  Our work goes further by
leveraging richer/fine-grained guidance which allows us to handle a larger
variety of tasks and addresses important sim2real aspects (such as removing explicit dependence
on reference trajectories from a motion model), enabling deployment on a real robot.


\section{The \modelname methodology for constructing humanoid skills}


Our approach starts conceptually with standard teleoperation RL pipelines for quadrupeds and bipeds / humanoids.  Typically such an RL policy (e.g.,~\cite{he2024learning,he2024omnih2o} is trained with a dense reward for accurately tracking keypoints from an input trajectory (obtained e.g., via motion capture of a human) along with other rewards (for stability, balance, smoothness, etc). However for training a humanoid to perform autonomous skills (such as picking up an object) an input trajectory is not available at test time so we need to either let RL learn the motion through exploration (which is very hard without careful reward engineering), or we need to generate this motion plan externally.  

In \modelname, we take this latter route of first generating motion plans externally through a pre-trained human motion prior.  These generated motion plans are then used implicitly during RL training in the reward signal but not explicitly used as goal conditions by the policy, (hence putting the ``Dream'' in “\modelname”).  In addition to these dense tracking rewards, we also use sparse and verifiable task-specific rewards to explicitly promote task completion. The overall pipeline is summarized in Fig.~\ref{fig:main}. We now discuss the two stages of (1) trajectory generation from a human motion prior and (2) RL in more detail. More details about the exact parameters used for each task in these 2 stages are in the Appendix.

\subsection{Stage 1: Generating Reference Trajectory from a Human Motion Prior}
A key desiderata for our first stage is to leverage human motion data instead of humanoid teleoperation data which is expensive.  Human motion data is widely available (either in the form of motion capture datasets or implicitly in video datasets) and thus allows us to learn high quality priors and a multitude of tasks.  By generating realistic human-like motions we also hope for more seamless sim-to-real transfer and also offer more natural interaction with humans.

Additionally, we would like to choose a model that has favorable scaling properties with respect to data.  We thus use a diffusion transformer~\cite{peebles2023scalable,tevet2023human,chi2023diffusion} which has been shown to be successful in modeling human motion as well as robot manipulation trajectories and is known to scale well with large datasets while remaining well behaved in low data regimes~\cite{prabhudesai2025diffusion}.

Among previous motion diffusion models, we build on OmniControl~\cite{xie2024omnicontrol} which can be flexibly conditioned on both text and spatiotemporal guidance.  OmniControl generates trajectories following a given text command (e.g., ``Pick up the bottle'') while stipulating that a joint or a subset of joints reach a prespecified spatial location at a prespecified time.  In this way, our trajectory generation stage is analogous to image or video inpainting.

This form of spatiotemporal guidance allows us to connect a generated trajectory to its environment (for example, allowing us to specify where a humanoid should sit, how high it should jump, or the location of the object to be manipulated).  Being able to control this interaction point is critical, since in the RL simulator we can instantiate an object at some location and then use trajectories that are guaranteed to approach this object, significantly facilitating the RL exploration problem. In~\modelname, we specifically design the form of spatiotemporal guidance for each task.
For instance, our pick task involves providing a spatial target for the wrist. See the Appendix for more details on guidance control implementations.

\parahead{Post-retargeting and trajectory filtering}
Since Omnicontrol is trained on human trajectories (represented via the SMPL parameterization~\cite{smpl}), we next retarget these generated trajectories to the G1 form factor (in similar fashion to \cite{pyroki2025}): by solving an optimization problem (using the  PyRoki~\cite{pyroki2025} library) that minimizes the relative keypoint positions, relative angles, and a scale factor that adjusts for the difference in link lengths. 
Additional residuals, such as feet contact costs, self-collision costs, and foot orientation costs, are used to improve physical plausibility.

Finally, we apply a layer of post-processing to the generated G1 trajectories prior to passing on to RL.  Some of the generated trajectories are not dynamically feasible and thus not fit to be used for tracking in Stage 2. We devise task-specific filtering mechanisms based on some heuristics as discussed in detail in the Appendix. We also apply task-specific trajectory refinements to avoid unnecessary movements, such as setting all left arm joints to a default value in the \texttt{Pick} task, where only right arm is used. These are also discussed in the Appendix.

\parahead{Trajectory representation}
After all post-filtering and refinement, we have a set of reference trajectories,
$\{\alpha_i\}$ generated with the same task-specific text prompt with 
different $(p_{k,i},\,t_{k,i})$ spatiotemporal ``goals'', which mean that
joint $k$ should be at position $p_{k,i}$ at
time $t_{k,i}$. In addition to this, we also define $t_g$ as the time at which the task-specific goal interaction occurs. For example, $t_g$ for the \texttt{Pick} task is the time when the object is to be picked up, $t_g$ for the \texttt{Button Press} task is when the button is to be pressed, etc.  This $t_g$ is crucial for synthesizing scenes for each given reference trajectory, as we discuss later in Section~\ref{sec:stage2}.

Each reference trajectory is represented as a sequence of target frames, 
$\alpha_i = [ \alpha_{i,0},\,\alpha_{i,\Delta t},...,\,\alpha_{i,(L-1)\Delta t}]$,
where $\Delta t=0.05s$ is the time step and $L=196$ is  trajectory length (hence each trajectory spans $9.8s$). Each frame is represented as
$\alpha_{t} = \{ p_t^{\text{ref,root}},\,\theta_t^{\text{ref,root}},\, q_t^{\text{ref}},\,s_t^{\text{ref,left}},\, s_t^{\text{ref,right}}\}$ where $q_t^{\text{ref}} \in \mathbb{R}^{27}$ are the reference joint angles, $p_t^{\text{ref,root}} \in \mathbb{R}^3$ is the position of the root, $\theta_t^{\text{ref,root}} \in \mathbb{R}^4$ is the orientation of the root in quaternions, and $s_t^{\text{ref,left}}, s_t^{\text{ref,right}}\in \{0, 1\}$ are the left and right hand states with $0$ denoting open and $1$ denoting closed. These are manually labeled for each task; for example, for the 
pick-with-right-hand task,
we ensure that the right hand closes
immediately after time $t_g$ and that the left hand stays closed through
the duration of the task.
Refer to the Appendix for more details.

\parahead{Out-of-distribution tasks}
We use OmniControl in this paper in a ``zero-shot'' fashion in the sense that we use the weights and hyperparameters as originally released by authors, and retarget them to G1 after trajectory generation.  Since OmniControl is trained on HumanML3d~\cite{Guo_2022_CVPR}, 
we find that it is capable of handling a wide variety of tasks ``out-of-the-box''.

However, we also explore a method handling certain novel tasks that are not well represented by Omnicontrol training distribution (e.g., pulling drawers) by using IK-based optimization on a base trajectory of a person standing idle (or bending down to pull a drawer below waist-level).
More details are described in the Appendix.



\subsection{Stage 2: RL with Reference Trajectory} \label{sec:stage2}
Once we have the reference trajectories from Stage 1, we formulate the interactive task as an RL problem.  In this section, we describe a
``privileged'' variant with access to internal simulator states 
and defer our discussion of how to adapt this approach for
real deployments to Section~\ref{sec:eval}.

\parahead{Scene synthesis}
First we need to synthesize a scene that makes sense for each of the Stage 1 kinematic trajectories to execute the interactive task --- 
for example if we had used guidance in Stage 1 to ask that the wrist
be at point $p$ at time $t$, then during RL we instantiate the object-to-be-manipulated near
point $p$.
More formally, given the time $t_g$ 
at which the interaction happens in the generated trajectory (e.g., the time at which the object is picked, button is pressed etc.), we place the object of interest (pick object, button etc.) at the following location:
\begin{align}
    &t^{\text{o,world}} = \textbf{t}_{t_g}^\text{b,world} + R_{t_g}^\text{b,world} \textbf{t}^\text{o,b}, \\
    &R^\text{o,world} = R_{t_g}^\text{b,world} R_{t_g}^\text{o,b},
\end{align}
where $(\textbf{t}_{t_g}^\text{b,world}, R_{t_g}^\text{b,world})$ is the pose of the robot body part link in question in world frame (e.g., the 
right wrist link for the pick-with-right-hand task) and
$(\textbf{t}^\text{o,b}, R^\text{o,b})$ is the offset of the object w.r.t the
robot body part link where the object should be placed. All specific offsets and body part links used in each task are listed in the Appendix.  We randomize the timestamp $t_g$, target positions used to generate trajectories, and thus $p_{t_g}$, and other characteristics of the object such as mass and friction.
We choose randomization hyperparameters to demonstrate a wide range of settings for which we can generate and use reference trajectories to solve tasks that can be scaled with careful engineering.
The exact randomization hyperparameters of the environment are reported in the Appendix. 

\parahead{Action space}
Next we define our action and observation spaces.
Our simulated robot is a 27-DoF Unitree G1 equipped with two 7-DoF DEX 3-1 hands, one mounted on each wrist (in real world experiments we use Inspire hands).
In this work we restrict hand control to 
discrete open/closed configurations that are fixed per-task (see Appendix for task-specific settings, e.g., extending the right index finger for the
open configuration during the button-press task). 
The action space is therefore defined as $a_t \in \mathbb{R}^{29}$, where $a_t = \{a_t^{\text{body}}, a_t^{\text{left}}, a_t^{\text{right}}\}$, with $a_t^{\text{body}} \in \mathbb{R}^{27}$ denoting the target joint angles for the G1 body, and $a_t^{\text{left}}, a_t^{\text{right}} \in \mathbb{R}$ controlling the left and right hands, respectively. For the hand controls, 
negative values correspond to an open hand and positive values to a closed hand. 



\begin{table*}[t!]
\centering
\caption{Reward terms for reference tracking and smooth policy enforcement.}
\begin{tabular}{ll}
\toprule
Reward Term & Interpretation \\
\midrule
$\lVert q_t^\text{robot} - q_t^\text{ref} \rVert_2$ & Penalizes deviation from reference joint angles \\
$\lVert p_{t}^{\text{ref,key}} - p_{t,\text{robot}}^{\text{key}} \rVert_2$ & Penalizes deviation from reference keypoints (3D positions in world frame) \\
$\lVert p_t^{\text{robot,root}} - p_t^{\text{ref,root}} \rVert$ & Penalizes deviation of robot root from reference root position \\
$|\theta_t^{\text{rel}}|$ & Penalizes deviation in orientation between robot and reference \\
$|\sigma(a_t^{\text{left}}) - s^{\text{ref, left}}| + |\sigma(a_t^{\text{right}}) - s_t^{\text{ref, right}}|$ & Penalizes deviation of hand states from reference ($\sigma(x)=\tfrac{1}{1+e^{-x}}$) \\
\arrayrulecolor{black!40}\midrule
$\lVert \tau_t \rVert_2 + \lVert \ddot{q}_t^\text{robot} \rVert_2$ & Penalizes high torques and accelerations \\
$\lVert \tfrac{a_t - a_{t-1}}{\Delta t} \rVert_2$ & Penalizes high action rate changes \\
$\sum_{k=\{\text{left foot}, \text{right foot}\}} \lVert c^{\text{f}} (pos_t^{\text{robot, f}} - pos_{t-1}^{\text{robot, f}})\rVert$ & Penalizes foot sliding while in ground contact \\
$n^{\text{feet}}$ & Penalizes excessive foot-ground contacts (to discourage baby steps) \\
$\theta_z^{\text{left foot,z}} + \theta_z^{\text{right foot,z}}$ & Encourages feet to remain parallel to the ground (discourages heel sliding) \\
\arrayrulecolor{black}\bottomrule
\end{tabular}
\label{tab:reward_terms}
\vspace{-0.1in}
\end{table*}

\parahead{Observations}
For each task, we include proprioception information (joint angles $q_t^\text{robot} \in \mathbb{R}^{27}$, joint velocities $\dot{q}_t^\text{robot} \in \mathbb{R}^{27}$), root linear velocity $v_t^{\text{root}} \in \mathbb{R}^3$, root angular velocity $\omega_t^{\text{root}} \in \mathbb{R}^3$, 
projected gravity in root frame $g_t \in \mathbb{R}^3$, previous action $a_{t-1} \in \mathbb{R}^{29}$, and a target trajectory reference as input observation, along with privileged task-specific observations like relative pose of the object, mass, friction of the object wherever relevant. At time $t$, the target trajectory reference observation consists of $[\gamma_t, \gamma_{t+\Delta t^\text{obs}},...,\gamma_{t+(K-1) \Delta t^\text{obs}}]$ where $K$ is the number of time steps 
into the future, and $\Delta t^\text{obs} = 0.1s$ is a hyperparameter. $\gamma_t$ consists of $(q_t^\text{ref},\dot{q}_t^\text{ref},p_t^{\text{rel, root}},p_t^{\text{rel, key}},s_t^{\text{ref, left}},s_t^{\text{ref, right}})$ where $q_t^\text{ref} \in \mathbb{R}^{27}$ are the target joint angles, $\dot{q}_t^\text{ref} \in \mathbb{R}^{27}$ are the target joint velocities, $p_t^{\text{rel,root}}$ is the relative pose of the root reference with respect to the robot's base, $p_t^{\text{rel, key}} \in \mathbb{R}^{3\times 41}$ corresponds to the relative position of the 41 keypoints on the robot with respect to it's root
and $s_t^{\text{ref,left}}, s_t^{\text{ref,right}}$ are the target reference binary hand states. Note that $\gamma_t$ essentially contains the same information as $\alpha_t$ but is transformed into the robot's frame and some redundant information is added for ease of policy learning inspired from \cite{chen2025gmt,he2024omnih2o,he2025asap}. Unlike these other works, we include relative pose $p_t^\text{rel, root}$ as input instead of root reference velocities, and target reference keypoints of reference with respect to robot's root instead of target trajectory's root. This is because works like \cite{chen2025gmt} do not aim to precisely track the trajectory but instead train a deployable policy that aims to follow velocity commands of root, and thus tend to drift from global reference trajectory. In our work, as we aim to precisely follow trajectories to accomplish interactive tasks, we exploit the privileged sim global root position to obtain relative keypoints as observation. However, it must be noted that it should be possible to later train a non-privileged vision-based policy that exploits scene information from vision to implicitly replace the global position privileged information.

\parahead{Rewards}
 In Table~\ref{tab:reward_terms}, we summarize our reward terms
(1) for tracking the reference correctly, (2)
to encourage maintaining balance and enable smooth control.

We also add some (3) task-specific rewards to encourage accomplishing the task with high success rates, such as the reward for raising an object above a height for the pick task. These are described for each task in the Appendix. The total reward $r_t$ is obtained by:\vspace{-2mm}
\[
r_t = \Sigma_{i=1}^{10} w_{r_i} r_{t,i} + w_{r^\text{task,sparse}} r^\text{task,sparse},\vspace{-2mm}
\]
where $w_{r_i} \text{ for } i \in \{1,2,...,10\}$ (indexing over our $10$ reward terms) and  $w_{r^\text{task,sparse}}$ are task-specific weights whose exact values are given in the Appendix.

\parahead{Training}
We setup our environment and training in
IsaacLab~\cite{mittal2023orbit} that uses IsaacSim simulation. All policies are trained on an NVIDIA RTX A6000 with $48$ GB vRAM using PPO~\cite{schulman2017proximal}. For each task, we train for $2000$ iterations with $8192$ parallel environments. See Appendix  for more details.

\section{Evaluation}\label{sec:eval}

\subsection{Tasks and Baselines}
We evaluate on a library of $11$ tasks: \texttt{Pick}, \texttt{Bimanual Pick},
\texttt{Pick from Ground (Side Grasp)}, \texttt{Pick from Ground (Top Grasp)},
\texttt{Press Button}, \texttt{Open Drawer}, \texttt{Open Door},
\texttt{Precise Punch},
\texttt{Precise Kick},
\texttt{Jump}, and \texttt{Sit}. For comparison, we report results for our method (\modelname), which combines tracking with task-specific sparse rewards, and three baselines: 
\begin{enumerate}[label=(\alph*)]
    \item \emph{TaskOnly}: only task-specific (sparse) rewards,
    \item \emph{TaskOnly+}: only task-specific rewards, both sparse and engineered dense rewards inspired by~\cite{luo2024omnigrasp}, and
    \item \emph{TrackingOnly}: only tracking rewards.
\end{enumerate}


\begin{table}[t!]
    \centering
    \caption{Success rates (\%) in simulation over $1000$ random environments. 
    (a) \emph{TaskOnly}; (b) \emph{TaskOnly+}; (c) \emph{TrackingOnly}. \\
    \textbf{Bold} denotes the best results.}
    \label{tab:results}
    \begin{tabular}{lcccc}
    \toprule
    Task / Method                 & (a) & (b) & (c) & Ours \\ 
    \midrule
    \texttt{Pick}                          & 0        & 15.1          & 87.5         & \textbf{95.4} \\ 
    \texttt{Bimanual Pick}                 & 0        & 31.0          & \textbf{100} & \textbf{100}  \\
    \texttt{Pick from Ground (Side Grasp)} & 0        & 0             & 99.4         & \textbf{100}  \\ 
    \texttt{Pick from Ground (Top Grasp)}  & 0        & 0             & \textbf{100} & \textbf{100}  \\ 
    \texttt{Press Button}                  & 0        & \textbf{99.8} & 99.1         & 99.3          \\ 
    \texttt{Open Drawer}                   & 0        & 24.5          & \textbf{100} & \textbf{100}  \\ 
    \texttt{Open Door}                     & 0        & 15.4          & \textbf{100} & \textbf{100}  \\ 
    \texttt{Precise Punch}                 & 0        & \textbf{100}  & 99.4         & 99.7          \\ 
    \texttt{Precise Kick}                  & 0        & 97.6          & 96.1         & \textbf{98.6} \\ 
    \texttt{Jump}                          & 0        & 0             & \textbf{100} & \textbf{100}  \\ 
    \texttt{Sit}                          & 0        & \textbf{100}             & \textbf{100} & \textbf{100}  \\ 
    \bottomrule
    \end{tabular}
\end{table}

\subsection{Simulation Results}
\begin{figure}[t!]
    \centering
    \includegraphics[
        trim={0 1.4in 12in 0}, 
        clip,
        width=0.95\columnwidth 
    ]{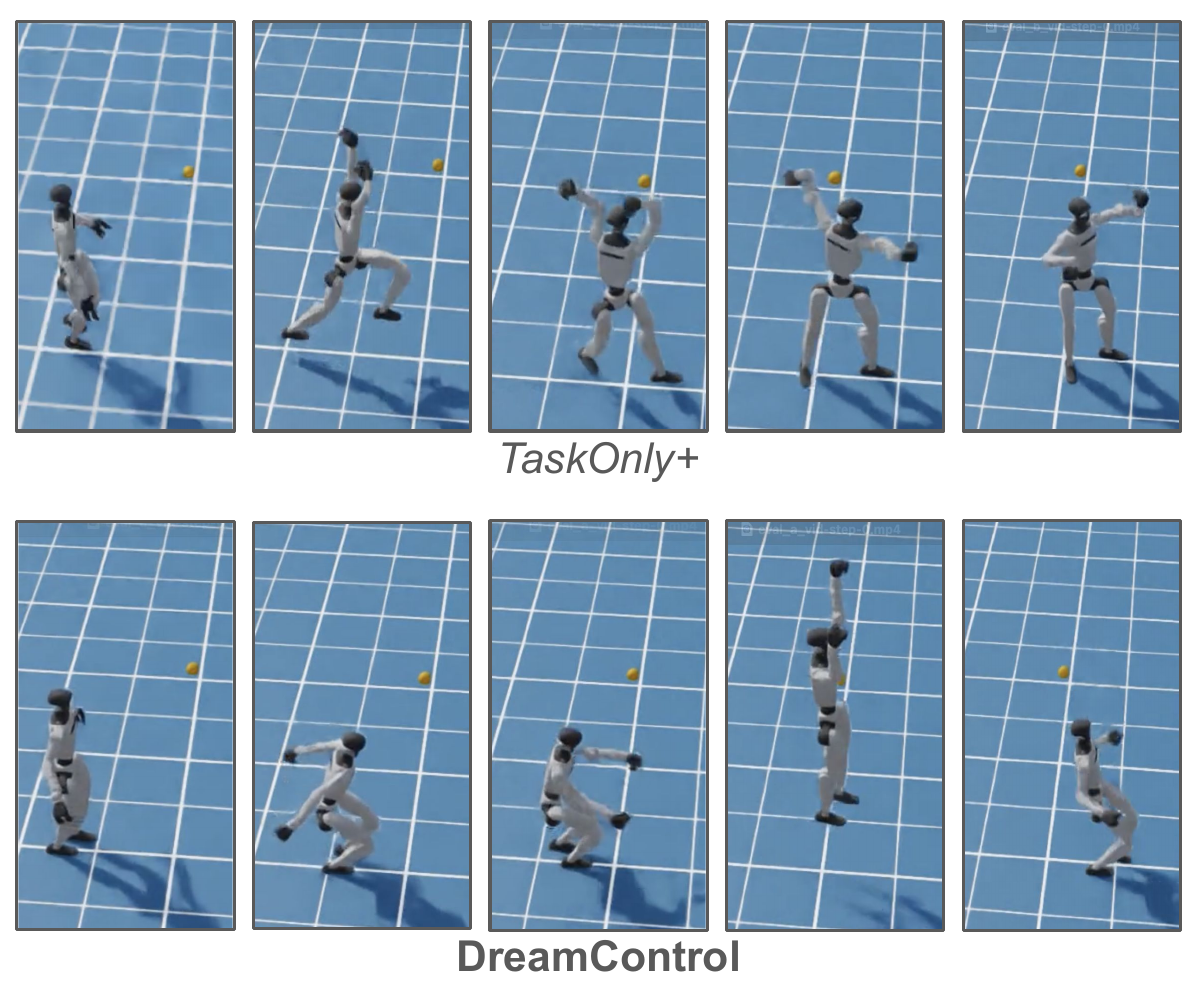}
    \caption{Comparison of trajectories for the task of \texttt{Jump}. The top row shows results from the \emph{TaskOnly+} baseline, while the bottom row illustrates trajectories from the \modelname. The yellow sphere depicts the spatial control point used to guide the trajectories.}
    \vspace{-3mm}
    \label{fig:jumping}
\end{figure}

Table~\ref{tab:results} reports success rates over $1000$ random environments, with each task success criteria defined in the Appendix. Our results show that
\emph{TaskOnly} (a) achieves $0\%$ success across all tasks, since relying solely on sparse rewards provides no dense guidance to discover meaningful motions. 
\emph{TaskOnly+} (b) improves performance by adding engineered dense terms, enabling success on simpler tasks like \texttt{Press Button} and \texttt{Precise Punch}, but still fails on tasks requiring coordinated whole-body motion. 
For example, in \texttt{Pick from Ground}, the robot must crouch in a balanced manner, and in \texttt{Jump} it must first lower its body before springing upward; with only a pelvis-target reward, the policy instead ``settles for'' merely stretching its knees but is unable to discover how to perform a real jump (see Fig.~\ref{fig:jumping}). 
\emph{TrackingOnly} (c)  performs better overall, but struggles with fine-grained interactive tasks such as \texttt{Pick}.
By combining both tracking and task-specific signals, \modelname~achieves robust performance, outperforming all baselines and achieving the best results on $9$ of $11$ tasks.

\subsection{Human-ness Comparison}
We also evaluate the human-ness (or naturalness) of trajectories generated by our policies compared to \emph{TaskOnly+}. 
For tasks where \emph{TaskOnly+} achieves non-zero success, we assess how natural the resulting motions appear. 
First, we report Fr\'echet inception distance (FID; \cite{heusel2017gans}) scores on the HumanML3D dataset~\cite{Guo_2022_CVPR}, using task-matched ground-truth trajectories obtained via keyword filtering (see Appendix ). 
As shown in Table~\ref{tab:humanness}, \modelname~consistently achieves lower FID values than \emph{TaskOnly+}, indicating closer alignment with human motions. 
An exception occurs in the \texttt{Pick} task, which we conjecture is due
to a domain gap: human demonstrations typically involve waist-level picking, whereas the shorter G1 robot often performs shoulder-level picks, making its trajectories less comparable to the human dataset. However, other metrics suggest that our method~\modelname produces more human-like motions consistently in all tasks.

To further assess motion quality, we calculate average absolute jerk (Table~\ref{tab:humanness}, second-to-last column), defined as $\frac{\sum_i \sum_t \sum_k |\dddot{p}_{t,k}^{\text{key,global}}|}{NTK}$, where $\dddot{p}_t^{\text{key,global}}$ is the third derivative of the global position of the $k^{\text{th}}$ keypoint, with $N=1000$ trajectories, $T=500$ time-steps, and $K=41$ keypoints. Lower jerk indicates smoother motions, and our method significantly outperforms \emph{TaskOnly+}, producing more fluid, human-like movements.  

We also conduct a user study with $40$ participants, in which they are shown side-by-side videos of trajectories from both methods (order randomized) and asked to select which looked more human-like. As summarized in Table~\ref{tab:humanness} (last column), participants overwhelmingly preferred \modelname's trajectories across all tasks, further confirming the naturalness of our approach.

Finally, as shown in Fig.~\ref{fig:jumping}, the trajectory generated by our approach (bottom row) is noticeably more natural and human-like compared to the \emph{TaskOnly+} baseline (top row). Our method exhibits a smooth jumping motion, where the robot first bends and then lifts off the ground, while the \emph{TaskOnly+} baseline lifts off but without bending, resulting in a less human-like motion that also does not accomplish the task.


\begin{table}
\centering
\caption{Human-ness comparison of \modelname~(Ours) and TaskOnly+. 
We report FID and jerk ($m/s^3$), where lower is better, and the average human preference.
\textbf{Bold} denotes the best results.}
\begin{tabular}{llccc}
\toprule
Task & Method & FID $\downarrow$ & Jerk $\downarrow$ & User Study $\uparrow$ \\
\midrule
\multirow{2}{*}{\texttt{Pick}}          
& \emph{TaskOnly+} & \textbf{0.240} & 211.2 & 15.0\% \\
& Ours & 0.320 & \textbf{147.5} & \textbf{85.0\%} \\                         
\arrayrulecolor{black!40}\midrule
\multirow{2}{*}{\texttt{Press Button}}
& \emph{TaskOnly+} & 1.220 & 235.7 & 17.25\% \\ 
& Ours & \textbf{0.375} & \textbf{161.9} & \textbf{82.75\%} \\  
\arrayrulecolor{black!40}\midrule
\multirow{2}{*}{\texttt{Precise Punch}}
& \emph{TaskOnly+} & 0.417 & 229.9 & 7.5\% \\ 
& Ours & \textbf{0.084} & \textbf{199.8} & \textbf{92.5\%} \\ 
\arrayrulecolor{black!40}\midrule
\multirow{2}{*}{\texttt{Precise Kick}}
& \emph{TaskOnly+} & 0.522 & 360.9 & 17.5\% \\ 
& Ours & \textbf{0.161} & \textbf{252.5} & \textbf{82.5\%} \\  
\arrayrulecolor{black!40}\midrule
\multirow{2}{*}{\texttt{Jump}}         
& \emph{TaskOnly+} & 1.216 & 236.4 & 5.0\% \\ 
& Ours & \textbf{0.208} & \textbf{148.5} & \textbf{95.0\%} \\  
\arrayrulecolor{black}\bottomrule
\end{tabular}\vspace{-4mm}
\label{tab:humanness}
\end{table}

\subsection{Sim2Real Deployment}
To demonstrate real-world effectiveness, we deploy policies for selected tasks on hardware after retraining with observations modified in the following ways 
to remove dependence on simulator-privileged information:
\begin{itemize}
    \item Remove the trajectory reference observation ($ref$), though references remain available via the rewards;
    \item Remove the linear velocity of the root;
    \item Remove privileged scene-physics information like object mass, friction etc.;
    \item Add time encoding $(t,\,\sin(2 \pi t / T))$, where $T$ is the total length of the episode.
\end{itemize}

We use the same rewards as in Stage~2, including motion tracking terms, but transform the reference trajectory for $(x, y, yaw)$ of the root to avoid privileged inputs that are unavailable for the critic.  
The resulting policy depends only on the relative position of the object / goal, making it deployable on the real robot.

\parahead{Hardware setup}  
We use a Unitree G1 humanoid ($27$-DoF, waist lock mode allowing only yaw movement) equipped with Inspire dexterous hands ($6$-DoF each, controlled in binary open / close mode). An onboard IMU provides root orientation, gravity direction, and angular velocity. A RealSense D435i depth camera, mounted on the neck, estimates the 3D position of the object / goal relative to the pelvis.

\parahead{Deployment}  
To estimate object positions, we leverage an off-the-shelf open-vocabulary object detection model, OWLv2~\cite{minderer2023scaling} for 2D localization, then lift to 3D using depth and object-specific offsets (Fig.~\ref{fig:main}(C)). Due to OWLv2’s inference latency, we detect the object only in the first frame and hold the estimate fixed thereafter. To mitigate errors from this static estimate, we freeze the lower body during interactive tasks (except bimanual pick) and add a penalty on root velocities to ensure the base remains static. Note that this limitation stems from perception bottlenecks rather than our method; in principle, vision-based policies could be trained via student-teacher distillation (e.g.,~\cite{lin2025sim}), which we leave for future work.
We provide details for sim2real transfer in the Appendix.
 
We successfully deployed policies for: \texttt{Pick} (standing), \texttt{Bimanual Pick} (with boxes of varying weights), \texttt{Press Button} (standing), \texttt{Open Drawer} (at different positions), \texttt{Precise Punch} (standing) and \texttt{Squat} (with varying depths). Representative visualizations are shown in Fig.~\ref{fig:splash} and more videos are available at \href{https://genrobo.github.io/DreamControl/}{Link}.

\vspace{-2mm}
\section{Discussion and Future Work}
In this work, we presented \modelname, a novel recipe for training autonomous humanoid skills that leverages guided diffusion for long-horizon planning and reinforcement learning for robust control, without requiring expensive demonstration data. We validated our approach on several challenging tasks, successfully transferring policies from simulation to a real G1 humanoid robot.

While our current implementation does not yet compose skills, support dexterous manipulation or complex object geometries, the data-efficient nature of our method provides a strong foundation for extensions along these lines.  
We believe that scaling \modelname~to a broader repertoire of tasks and more diverse robot morphologies is a promising and immediate next step toward more capable and general-purpose humanoid robots.





\section*{ACKNOWLEDGMENT}
We are thankful for discussions and help with 
experiments from B. Rishi, Y. Patel, D. Narayanan, S. Deolasee, V. Rajesh, G. Moffatt.


\bibliographystyle{IEEEtran}
\bibliography{IEEEabrv,references}

\clearpage
\section*{APPENDIX}


\section{Robot details}

\subsection{Joints}

Our Unitree G1 Edu+ consists of $27$ joints with their names as mentioned in Table \ref{tab:joints_grouped}.

\begin{table}[h]
\centering
\caption{Joints of the Unitree G1 Edu+ grouped by body part.}
\label{tab:joints_grouped}
\begin{tabular}{ll}
\toprule
\textbf{Legs} & \\
\arrayrulecolor{black!40}\midrule
left\_hip\_pitch\_joint & right\_hip\_pitch\_joint \\
left\_hip\_roll\_joint  & right\_hip\_roll\_joint  \\
left\_hip\_yaw\_joint   & right\_hip\_yaw\_joint   \\
left\_knee\_joint       & right\_knee\_joint       \\
left\_ankle\_pitch\_joint & right\_ankle\_pitch\_joint \\
left\_ankle\_roll\_joint  & right\_ankle\_roll\_joint  \\
\arrayrulecolor{black}\midrule
\textbf{Waist} & \\
\arrayrulecolor{black!40}\midrule
waist\_yaw\_joint & \\
\arrayrulecolor{black}\midrule
\textbf{(Left $|$ Right) Arms} & \\
\arrayrulecolor{black!40}\midrule
left\_shoulder\_pitch\_joint & right\_shoulder\_pitch\_joint \\
left\_shoulder\_roll\_joint  & right\_shoulder\_roll\_joint  \\
left\_shoulder\_yaw\_joint   & right\_shoulder\_yaw\_joint   \\
left\_elbow\_joint           & right\_elbow\_joint \\
left\_wrist\_roll\_joint     & right\_wrist\_roll\_joint \\
left\_wrist\_pitch\_joint    & right\_wrist\_pitch\_joint \\
left\_wrist\_yaw\_joint      & right\_wrist\_yaw\_joint \\
\arrayrulecolor{black}\midrule
\textbf{(Left $|$ Right) Hands} & \\
\arrayrulecolor{black!40}\midrule
left\_hand\_index\_0\_joint  & right\_hand\_index\_0\_joint \\
left\_hand\_index\_1\_joint  & right\_hand\_index\_1\_joint  \\
left\_hand\_middle\_0\_joint & right\_hand\_middle\_0\_joint \\
left\_hand\_middle\_1\_joint & right\_hand\_middle\_1\_joint  \\
left\_hand\_thumb\_0\_joint  & right\_hand\_thumb\_0\_joint   \\
left\_hand\_thumb\_1\_joint  & right\_hand\_thumb\_1\_joint \\
left\_hand\_thumb\_2\_joint  & right\_hand\_thumb\_2\_joint \\
\arrayrulecolor{black}\bottomrule
\end{tabular}
\end{table}

\subsection{Keypoints}

The $41$ keypoints on the robot are named as detailed in Table \ref{tab:keypoints_grouped}.

\begin{table}[h]
    \centering
    \caption{Keypoints of the Unitree G1 Edu+ grouped by body part.}
    \label{tab:keypoints_grouped}
    \begin{tabular}{p{0.45\linewidth} p{0.45\linewidth}}
        \toprule
        \textbf{Legs} & \\
        \arrayrulecolor{black!40}\midrule
        left\_hip\_pitch\_link & right\_hip\_pitch\_link \\
        left\_hip\_roll\_link  & right\_hip\_roll\_link  \\
        left\_hip\_yaw\_link   & right\_hip\_yaw\_link   \\
        left\_knee\_link       & right\_knee\_link       \\
        left\_ankle\_pitch\_link & right\_ankle\_pitch\_link \\
        left\_ankle\_roll\_link  & right\_ankle\_roll\_link  \\
        \arrayrulecolor{black}\midrule
        \textbf{Waist \& Torso} & \\
        \arrayrulecolor{black!40}\midrule
        pelvis & pelvis\_contour\_link \\
        waist\_yaw\_link & waist\_roll\_link \\
        torso\_link & waist\_support\_link \\
        logo\_link & \\
        \arrayrulecolor{black}\midrule
        \textbf{Head \& Sensors} & \\
        \arrayrulecolor{black!40}\midrule
        head\_link & imu\_link \\
        d435\_link & mid360\_link \\
        \arrayrulecolor{black}\midrule
        \textbf{Arms} & \\
        \arrayrulecolor{black!40}\midrule
        left\_shoulder\_pitch\_link & right\_shoulder\_pitch\_link \\
        left\_shoulder\_roll\_link  & right\_shoulder\_roll\_link  \\
        left\_shoulder\_yaw\_link   & right\_shoulder\_yaw\_link   \\
        left\_elbow\_link           & right\_elbow\_link \\
        left\_wrist\_roll\_link     & right\_wrist\_roll\_link \\
        left\_wrist\_pitch\_link    & right\_wrist\_pitch\_link \\
        left\_wrist\_yaw\_link      & right\_wrist\_yaw\_link \\
        left\_rubber\_hand          & right\_rubber\_hand \\
        \arrayrulecolor{black}\bottomrule
    \end{tabular}
\end{table}

\subsection{PD control}

We use PD controller to convert target joint angles to torque, $\tau_t$ as follows:-
\begin{equation}
    \tau_t = k_p (q_t^\text{commands} - q_t^\text{robot}) - k_d \dot{q}_t^\text{robot} \nonumber
\end{equation}

Where $k_p$ and $k_d$ are positional and derivative gains and $q_t^\text{commands}$ are the joint commands. The $k_p, k_d$ gains for each joint are given in Table \ref{tab:joint_gains_columnwise}

\begin{table}[h]
\centering
\caption{Joint list (unrolled column-wise) with default angle, $K_p$, and $K_d$ all initialized to $0$.}
\label{tab:joint_gains_columnwise}
\begin{tabular}{lccc}
\toprule
\textbf{Joint name} & \textbf{Default angle} & \textbf{$K_p$} & \textbf{$K_d$} \\
\midrule
left\_hip\_pitch\_joint   & -0.2 & 200 & 5 \\
left\_hip\_roll\_joint    & 0 & 150 & 5 \\
left\_hip\_yaw\_joint     & 0 & 150 & 5 \\
left\_knee\_joint         & 0.42 & 200 & 5 \\
left\_ankle\_pitch\_joint & -0.23 & 20 & 2 \\
left\_ankle\_roll\_joint  & 0 & 20 & 2 \\
right\_hip\_pitch\_joint     & -0.2 & 200 & 5 \\
right\_hip\_roll\_joint      & 0 & 150 & 5 \\
right\_hip\_yaw\_joint       & 0 & 150 & 5 \\
right\_knee\_joint           & 0.42 & 200 & 5 \\
right\_ankle\_pitch\_joint   & -0.23 & 20 & 2 \\
right\_ankle\_roll\_joint    & 0 & 20 & 2 \\
waist\_yaw\_joint            & 0 & 200 & 5 \\
left\_shoulder\_pitch\_joint & 0.35 & 40 & 10 \\
left\_shoulder\_roll\_joint  & 0.16 & 40 & 10 \\
left\_shoulder\_yaw\_joint   & 0 & 40 & 10 \\
left\_elbow\_joint           & 0.87 & 40 & 10 \\
left\_wrist\_roll\_joint     & 0 & 40 & 10 \\
left\_wrist\_pitch\_joint    & 0 & 40 & 10 \\
left\_wrist\_yaw\_joint      & 0 & 40 & 10 \\
left\_hand\_index\_0\_joint  & 0 & 5 & 1.25 \\
left\_hand\_index\_1\_joint  & 0 & 5 & 1.25 \\
left\_hand\_middle\_0\_joint & 0 & 5 & 1.25 \\
left\_hand\_middle\_1\_joint & 0 & 5 & 1.25 \\
left\_hand\_thumb\_0\_joint  & 0 & 5 & 1.25 \\
left\_hand\_thumb\_1\_joint  & 0 & 5 & 1.25 \\
left\_hand\_thumb\_2\_joint  & 0 & 5 & 1.25 \\
right\_shoulder\_pitch\_joint& 0.35 & 40 & 10 \\
right\_shoulder\_roll\_joint & -0.16 & 40 & 10 \\
right\_shoulder\_yaw\_joint  & 0 & 40 & 10 \\
right\_elbow\_joint          & 0.87 & 40 & 10 \\
right\_wrist\_roll\_joint    & 0 & 40 & 10 \\
right\_wrist\_pitch\_joint   & 0 & 40 & 10 \\
right\_wrist\_yaw\_joint     & 0 & 40 & 10 \\
right\_hand\_index\_0\_joint & 0 & 5 & 1.25 \\
right\_hand\_index\_1\_joint & 0 & 5 & 1.25 \\
right\_hand\_middle\_0\_joint& 0 & 5 & 1.25 \\
right\_hand\_middle\_1\_joint& 0 & 5 & 1.25 \\
right\_hand\_thumb\_0\_joint & 0 & 5 & 1.25 \\
right\_hand\_thumb\_1\_joint & 0 & 5 & 1.25 \\
right\_hand\_thumb\_2\_joint & 0 & 5 & 1.25 \\
\bottomrule
\end{tabular}
\end{table}

\section{Open and closed states for each task}

Each task has a specific set of joint angles for open and closed hand states to facilitate task-specific function. For instance, we want all fingers to open and closed in open and closed states of hand respectively while for button pressing, we only need the index finger open. For boxing, we keep all fingers closed for both open and closed as we never need to open them for the task. The exact finger joint angles for open and closed hand states for each task are listed in Table \ref{tab:hand_states}.

\begin{table}[h]
\centering
\caption{Open and Closed hand states for all tasks. Each config state for a hand consists of a tuple of $7$ joint angles in the same order as in Table \ref{tab:body_joints_grouped}. AOL: $(0,0,0,0,0,0,0)$, ACL: $(-\pi/2,-\pi/2,-\pi/2,-\pi/2,0,\pi/3,\pi/2)$, AOR: $(0,0,0,0,0,0,0)$, ACR: $(\pi/2,\pi/2,\pi/2,\pi/2,0,\pi/3,\pi/2)$, BPR: $(0,0,\pi/2,\pi/2,0,\pi/3,\pi/2)$, DOR: $(0,\pi/2,0,\pi/2,0,0,0)$}
\label{tab:hand_states}
\begin{tabular}{lccc}
\toprule
Task & \begin{tabular}[c]{@{}c@{}@{}}Left Hand \\ Config\\ (Open | Close)\end{tabular} 
     & \begin{tabular}[c]{@{}c@{}@{}}Right Hand \\ Config\\ (Open | Close)\end{tabular} 
     & \begin{tabular}[c]{@{}c@{}@{}}Group \\set to \\ default $q$\\ \end{tabular} \\ 
\midrule
\texttt{Pick}                                                                     & ACL | ACL & AOR | ACR & $G^\text{left arm}$\\ 
\texttt{Precise Punch}                                                            & ACL | ACL & ACR | ACR & - \\ 
\texttt{Precise Kick}                                                             & ACL | ACL & ACR | ACR & - \\ 
\texttt{Press Button}                                                             & ACL | ACL & PBR | PBR & - \\ 
\texttt{Jump}                                                                     & ACL | ACL & ACR | ACR & - \\ 
\texttt{Sit}                                                                      & ACL | ACL & ACR | ACR & - \\ 
\texttt{Bimanual} \texttt{Pick}                                                            & ACL | ACL & ACR | ACR & - \\ 
\begin{tabular}[c]{@{}l@{}}\texttt{Pick from Ground} \\ \texttt{(side grasp)}\end{tabular} & AOL | ACL & ACR | ACR & $G^\text{right arm}$\\ 
\begin{tabular}[c]{@{}l@{}}\texttt{Pick from Ground} \\ \texttt{(top grasp)}\end{tabular}  & ACL | ACL & ACR | ACR & $G^\text{left arm}$ \\ 
\texttt{Pick} \texttt{and Place}                                                           & ACL | ACL & AOR | ACR & - \\ 
\texttt{Open Drawer}                                                              & ACL | ACL & DOR | ACR & - \\ 
\texttt{Open Door}                                                                & ACL | ACL & DOR | ACR & - \\ 
\bottomrule
\end{tabular}
\end{table}

\section{Keyword-based filtering for FID}

We use the following keywords to filter out trajectories that are used for FID calculation to evaluate human-ness of keypoint trajectories followed by \textbf{Ours} against \emph{TaskOnly+}:-

\begin{itemize}
    \item \texttt{Pick}: All motions whose prompts contain ``pick" keyword in them
    \item \texttt{Press Button}: All motions whose prompts contain ``press" keyword and ``button" keyword in them
    \item \texttt{Precise Punch}: All motions whose prompts contain ``punch" keyword in them
    \item \texttt{Precise Kick}: All motions whose prompts contain ``kick" keyword in them
    \item \texttt{Jump}: All motions whose prompts contain ``jump" keyword in them   
\end{itemize}

\section{Reference trajectory generation}

For each task, we have a text prompt, $\lambda^\text{text}$ and a spatial control signal $\lambda^\text{spatial} \in \mathcal{R}^{L * S * 3}$ where $L=196$ is the no of time-step, $S=22$ is the number of SMPL joints named as follows:- 

\begin{table}[H]
    \centering
    \caption{Body joints grouped by body part.}
    \label{tab:body_joints_grouped}
    \begin{tabular}{p{0.45\linewidth} p{0.45\linewidth}}
        \toprule
        \textbf{Legs} & \\
        \arrayrulecolor{black!40}\midrule
        left\_hip   & right\_hip \\
        left\_knee  & right\_knee \\
        left\_ankle & right\_ankle \\
        left\_foot  & right\_foot \\
        \arrayrulecolor{black}\midrule
        \textbf{Spine \& Torso} & \\
        \arrayrulecolor{black!40}\midrule
        pelvis & spine\_1 \\
        spine\_2 & spine\_3 \\
        neck & head \\
        \arrayrulecolor{black}\midrule
        \textbf{Arms} & \\
        \arrayrulecolor{black!40}\midrule
        left\_collar    & right\_collar \\
        left\_shoulder  & right\_shoulder \\
        left\_elbow     & right\_elbow \\
        left\_wrist     & right\_wrist \\
        \arrayrulecolor{black}\bottomrule
    \end{tabular}
\end{table}

For a given spatial control signal point at a given time-step and joint, it is considered functional only if the 3D spatial point is not $(0, 0, 0)$. Hence, we initiate all spatial control points with $(0, 0, 0)$ and then fill out values of joints at time-steps that we want to control.

\subsection{Task-specific prompts}

All task-specific prompt texts, $\lambda^\text{text}$ are listed in Table \ref{tab:text_grompts}. The spatial control signals for each task are given as follows:-

\begin{table*}[t!]
{\centering
\caption{Text prompts for each task in simulation}
\begin{tabular}{ll}
\toprule
Task   & Prompts   \\ 
\midrule
\texttt{Pick}                            & ``a person walks to cup, grabs the cup from side and lifts up''            \\ 
\texttt{Precise Punch}                   & ``a person performs a single boxing punch with his right hand''            \\ 
\texttt{Precise Kick}                    & ``a person stands and kicks with his right leg''                           \\ 
\texttt{Press Button}                    & ``a person walks towards elevator, presses elevator button''               \\ 
\texttt{Jump}                            & ``a person jumps forward''                                                 \\ 
\texttt{Sit}                             & ``a person walks towards a chair, sits down''                              \\ 
\texttt{Bimanual} \texttt{Pick}                   & ``a person raises the toolbox with both hands''                            \\ 
\texttt{Pick} \texttt{from Ground (Side Grasp)}   & ``a person raises the toolbox with the use of one hand''                   \\ 
\texttt{Pick} from. Ground (Top Grasp)   & ``a person walks forward, bends down to pick something up off the ground'' \\ 
\texttt{Pick} \texttt{and Place}                  & ``a person picks the cup and puts it on another table''                    \\ \bottomrule
\end{tabular}

\label{tab:text_grompts}
}
\end{table*}

\begin{table*}[t!]
{\centering
\caption{Text prompts and slow-down factor for each task deployed on real robot}
\begin{tabular}{llc}
\toprule
Task   & Prompts  & Slow down factor \\ 
\midrule
\texttt{Pick}                            & ``a person stands in place, grabs the cup from side and lifts up''  & 2.5          \\ 
\texttt{Precise Punch}                   & ``a person performs a single boxing punch with his right hand''    & 1.5        \\ 
\texttt{Press Button}                    & ``a person stands in place, presses elevator button''        & 1.5       \\ 
\texttt{Bimanual} \texttt{Pick}                   & ``a person raises the toolbox with both hands''         & 1                   \\ 
\texttt{Squat}   & ``a person squats in place and stands up'' & 1                   \\  \bottomrule
\end{tabular}

\label{tab:text_prompts_real}
}
\end{table*}

\subsubsection{\texttt{Pick}}

For each reference trajectory:- We sample a target point, $p^\text{target} = \{p^x \in \mathcal{U}(1.0,1.2), p^y \in \mathcal{U}(-0.4,0.4),p^z=1.1 \}$. Target time-step is chosen to be, $t_g'=50+\lfloor 50 (p^x-1.0)\rfloor$. Then, we set spatial control signal for wrist as follows:-
\begin{align*} \notag
    & \lambda^\text{right\_wrist}_i = p^\text{target} \forall i \in \{t_g',...,t_g'+20\} \nonumber \\
    & \lambda^\text{right\_wrist}_i = (p^\text{target,x},p^\text{target,y},p^\text{target,z}+0.2) \\
    & \ \ \ \ \ \ \ \ \ \ \ \ \ \ \forall i \in \{t_g'+40,...,L\} \nonumber \\
\end{align*}

We also set the target spatial points for elbow to encourage generating trajectories where the object is grabbed from side as follows:-
\begin{align*}
    \lambda^\text{elbow}_i = &(p^\text{target,x},p^\text{target,y}-0.26\cos(\frac{\pi}{4}),p^\text{target,z}-0.26\sin(\frac{\pi}{4})) \nonumber\\
    &\forall i \in \{t_g',...,t_g'+20\} \nonumber\\
\end{align*}

\subsubsection{\texttt{Precise Punch}}

For each reference trajectory:- We sample a target point, $p^\text{target} = \{p^x \in \mathcal{U}(1.2,1.5), p^y \in \mathcal{U}(-0.2,0.0),p^z=\mathcal{U}(1.0,1.5) \}$. Target time-step is chosen to be, $t_g'=30$. Then, we set spatial control signal for wrist as follows:-
\begin{align*}
    & \lambda^\text{right\_wrist}_i = p^\text{target} \forall i \in \{t_g'-10,...,t_g'+10\} \nonumber\\
\end{align*}

\subsubsection{\texttt{Precise Kick}}

For each reference trajectory:- We sample a target point, $p^\text{target} = \{p^x \in \mathcal{U}(1.0,1.2), p^y = 0.0,p^z=\mathcal{U}(0.5,1.0) \}$. Target time-step is chosen to be, $t_g'=30$. Then, we set spatial control signal for right foot as follows:-
\begin{align*}
    & \lambda^\text{right\_foot}_i = p^\text{target} \forall i \in \{t_g'-13,...,t_g'+10\} \nonumber\\
\end{align*}

\subsubsection{\texttt{Press Button}}

For each reference trajectory:- We sample a target point, $p^\text{target} = \{p^x \in \mathcal{U}(1.4,1.8), p^y = \mathcal{U}(-0.4,0.4),p^z=\mathcal{U}(1.1,1.2) \}$. Target time-step is chosen to be, $t_g'=70+\lfloor 50 (p^x-1.4)\rfloor$. Then, we set spatial control signal for right foot as follows:-
\begin{align*}
    & \lambda^\text{right\_wrist}_i = p^\text{target} \forall i \in \{t_g',...,t_g'+20\} \nonumber\\
    & \lambda^\text{right\_wrist}_i = {p^\text{target,x}-0.2,p^\text{target,y},p^\text{target,z}} \text{ for } i = t_g'+40 \nonumber\\
\end{align*}

\subsubsection{\texttt{Jump}}

We sample a trajectory with target point, $p^\text{target} = \{p^x=1.0,p^y=0.0,p^z=1.9\}$. Target time-step is chosen to be $t_g'=50$. We set the spatial control signal for pelvis as: $\lambda^\text{pelvis}_i = p^\text{target} \text{ for } i=t_g'$

\subsubsection{\texttt{Sit}}

We sample a trajectory with target point, $p^\text{target} = \{p^x=1.3,p^y=0.8,p^z=0.58\}$. Target time-step is chosen to be $t_g'=100$. We set the spatial control signal for pelvis as: $\lambda^\text{pelvis}_i = p^\text{target} \text{ for } i=t_g'$

\subsubsection{\texttt{Bimanual} \texttt{Pick}}

We use target point, $p^\text{target} = \{p^x=0.7,p^y=0.0,p^z=0.65\}$. Target time-step is chosen to be $t_g'=98$. We set the spatial control signal for left wrist as follows:- 

\begin{align*}
    & \lambda^\text{left\_wrist}_i = \{p^\text{target,x},p^\text{target,y}+0.35,p^\text{target,z}+0.25\\
    & \ \ \ \ \ \ \ \ \ \ \ \ \ \ \ -\frac{0.25i}{t_g'}\} \forall i \in \{0,...,t_g'\} \nonumber \\
    & \lambda^\text{left\_wrist}_i = \{p^\text{target,x},p^\text{target,y}+0.15,p^\text{target,z}-0.25\\
    &\ \ \ \ \ \ \ \ \ \ \ \ \ \ \ \ +\frac{0.25i}{t_g'}\} \forall i \in \{t_g',...,t_g'+98\} \nonumber \\
\end{align*}

Spatial control signal for right wrist is as follows:-

\begin{align*}
    & \lambda^\text{right\_wrist}_i = \{p^\text{target,x},p^\text{target,y}-0.35,p^\text{target,z}+0.25\\
    &\ \ \ \ \ \ \ \ \ \ \ \ \ \ \ \ -\frac{0.25i}{t_g'}\} \forall i \in \{0,...,t_g'\} \nonumber \\
    & \lambda^\text{right\_wrist}_i = \{p^\text{target,x},p^\text{target,y}-0.15,p^\text{target,z}-0.25\\
    &\ \ \ \ \ \ \ \ \ \ \ \ \ \ \ \ +\frac{0.25i}{t_g'}\} \forall i \in \{t_g',...,t_g'+98\} \nonumber \\
\end{align*}

\subsubsection{\texttt{Pick} \texttt{from Ground (Side Grasp)}}

We use target point, $p^\text{target} = \{p^x=0.5,p^y=0.5,p^z=0.1\}$. Target time-step is $t_g'=98$. We set the spatial control signal for the left wrist as follows:-

\begin{align*}
    & \lambda^\text{left\_wrist}_i = \{p^\text{target,x},p^\text{target,y},p^\text{target,z}+0.5-\frac{0.5i}{t_g'}\} \\
    &\ \ \ \ \ \ \ \ \ \ \ \ \ \ \ \ \forall i \in \{0,...,t_g'\} \nonumber \\
    & \lambda^\text{left\_wrist}_i = \{p^\text{target,x},p^\text{target,y},p^\text{target,z}-0.5+\frac{0.5i}{t_g'}\} \\
    &\ \ \ \ \ \ \ \ \ \ \ \ \ \ \ \ \forall i \in \{t_g',...,t_g'+98\} \nonumber \\
\end{align*}

\subsubsection{\texttt{Pick} \texttt{from Ground (Top Grasp)}}

We use target point, $p^\text{target} = \{p^x=1.0,p^y=0.0,p^z=0.2\}$. Target time-step is $t_g'=50$. We set the spatial control signal for the right wrist as follows:-

\begin{align*}
    & \lambda^\text{right\_wrist}_i = \{p^\text{target,x},p^\text{target,y},p^\text{target,z}\} \text{ for } i=t_g' \nonumber \\
\end{align*}

\subsubsection{\texttt{Pick} \texttt{and Place}}

We use 2 target points, $p^\text{target,1} = \{p^x=1.2,p^y=-0.15,p^z=0.6\}$ and $p^\text{target,2} = \{p^x=1.2,p^y=-0.6,p^z=0.6\}$. Target time-steps are $t_{g1}'=80$ and $t_{g2}'=160$. We set the spatial control signal for the right wrist as follows:-

\begin{align*}
    & \lambda^\text{right\_wrist}_i = p^\text{target,1} \text{ for } i=t_{g1}' \nonumber \\
    & \lambda^\text{right\_wrist}_i = p^\text{target,2} \text{ for } i=t_{g2}' \nonumber \\
\end{align*}

\subsubsection{\texttt{Open Drawer}}

For drawer opening, the number of trajectories in HumanML dataset are very limited (only $18$ with both ``drawer" and ``open" keywords in text description), due to which the generated motion was very off when prompted with drawer opening and only giving sparse spatial control signal for wrist at the drawer. To address this, we prompt \modelname~to generate $2$ trajectories and no spatial control signal, one with prompt ``stand still" and one with prompt ``squat and stay in squat position". We use the generated motion as initialization and solve for the following optimization problem with gradient descent from the current state. The target points are sampled as $p^\text{target}=\{p^x \in \mathcal{U}(0.3, 0.35), p^y \in \mathcal{U}(-0.2,-0.1),p^z \in \mathcal{U}(0.4,0.8)\}$ and $t_g'=40$ if $p^z \ge 0.7$ and $t_g'=50$ if $p^z < 0.7$.

We define target trajectory, $\tau^\text{wrist}$ for wrist as follows if $p^z>0.7$:-

\begin{align*}
    \tau_i^\text{wrist} &= \{(p^\text{target,x})*\frac{i}{40},-0.25+(p^\text{target,y}+0.25)*\frac{i}{40},\\
    &\ \ \ \ \ \ \ \ \ \ \ \ \ \ \ \ 0.7+(p^\text{target,z}-0.7)*quad(\frac{i}{40})\} \nonumber \\
    &\text{where } quad(x) = 1 - (x-1)^2, \forall i \in \{0,1...40\}  \nonumber \\
    \tau_i^\text{wrist} &= \{p^\text{target,x}, p^\text{target,y},p^\text{target,z}\} \forall i \in \{41,42...70\} \nonumber \\
    \tau_i^\text{wrist} &= \{p^\text{target,x} - \frac{0.2(i-70)}{40},p^\text{target,y},p^\text{target,z}\} \\
    &\ \ \ \ \ \ \ \ \ \ \ \ \ \ \ \ \forall i \in \{71,72...110\} \nonumber \\
    \tau_i^\text{wrist} &= \{p^\text{target,x} - 0.2,p^\text{target,y},p^\text{target,z}\} \forall i \in \{111,112...196\} \nonumber \\
\end{align*}

and for $p^z<0.7$ where $w_i$ is the wrist position in the original re-targeted trajectory.
\begin{align*}
    \tau_i^\text{wrist} &= w_i \forall i \in \{0,1,...,50\} \nonumber \\
    \tau_{i+50}^\text{wrist} &= \{w_{50}^x+(p^\text{target,x}-w_{50}^x)*\frac{i}{40},w_{50}^y+(p^\text{target,y}-\\
    &\ \ \ \ \ \  w_{50}^y)*\frac{i}{40},w_{50}^z+(p^\text{target,z}-w_{50}^z)*quad(\frac{i}{40})\} \nonumber \\
    &\text{where } quad(x) = 1 - (x-1)^2, \forall i \in \{0,1,...,40\} \nonumber \\
    \tau_{i+50}^\text{wrist} &= \{p^\text{target,x}, p^\text{target,y},p^\text{target,z}\} \forall i \in \{41,42...70\} \nonumber \\
    \tau_{i+50}^\text{wrist} &= \{p^\text{target,x} - \frac{0.2(i-70)}{40},p^\text{target,y},p^\text{target,z}\} \\
    &\ \ \ \ \ \ \ \ \ \ \ \ \ \ \ \ \forall i \in \{71,72,...,110\} \nonumber \\
    \tau_{i+50}^\text{wrist} &= \{p^\text{target,x} - 0.2,p^\text{target,y},p^\text{target,z}\} \\
    &\ \ \ \ \ \ \ \ \ \ \ \ \ \ \ \ \forall i \in \{111,112,...,146\} \nonumber \\
\end{align*}

With this target trajectory, we run gradient descent to optimize only the joint angles $q_i^\text{ref}$ with the following loss in order to align the wrist to the target trajectory:-

\begin{align*}
    L &= \Sigma_{i=0}^{i=L-1} (p_i^\text{wrist} - w_i)^2 \nonumber \\
    &\text{where } p_i^\text{wrist} = fk^\text{wrist}(p_i^\text{ref,root},\theta_i^\text{ref,root},q_i^\text{ref,*}) \nonumber \\
    &q_i^\text{ref,j+1} = q_i^\text{ref,j} - \lambda^\text{lr} \frac{\partial}{\partial q_i^\text{ref,*}} L \nonumber
\end{align*}

\subsubsection{\texttt{Open Door}}

We generate trajectories for door opening in a similar way to drawer opening with the only changes in the target trajectory as follows for $p^\text{target,z}\ge0.7$:-

\begin{align*}
    \tau_i^\text{wrist} &= \{(p^\text{target,x})*\frac{i}{40},-0.25+(p^\text{target,y}+0.25)*\frac{i}{40},\\
    &\ \ \ \ \ \ \ \ \ \ \ \ \ \ \ \ 0.7+(p^\text{target,z}-0.7)*quad(\frac{i}{40})\} \nonumber \\
    &\text{where } quad(x) = 1 - (x-1)^2, \forall i \in \{0,1...40\} \nonumber \\
    \tau_i^\text{wrist} &= \{p^\text{target,x}, p^\text{target,y},p^\text{target,z}\} \forall i \in \{41,42...70\} \nonumber \\
    \tau_i^\text{wrist} &= \{c^x - R \sin(a_i) - h \cos(a_i),c^y + R \cos(a_i) \\
    &\ \ \ \ \ \ \ \ \ \ \ \ \ \ \ \ - h \sin(a_i),p^\text{target,z}\} \nonumber \\
    &\forall i \in \{71,72...110\} \text{ where } c^x = p^\text{target,x} + h, \\
    &\ \ \ \ \ \ \ \ \ \ \ \ \ \ \ \ c^y = p^\text{target,y}-R, a_i = \frac{i-70}{40} \nonumber \\
    \tau_i^\text{wrist} &= \{p^\text{target,x} - 0.2,p^\text{target,y},p^\text{target,z}\} \forall i \in \{111,112...196\} \nonumber \\
\end{align*}

and for $p^\text{target,z} < 0.7$ :-

\begin{align*}
    \tau_i^\text{wrist} &= w_i \forall i \in \{0,1,...,50\} \nonumber \\
    \tau_{i+50}^\text{wrist} &= \{w_{50}^x+(p^\text{target,x}-w_{50}^x)*\frac{i}{40},w_{50}^y+(p^\text{target,y}\\
    &\ \ \ \ -w_{50}^y)*\frac{i}{40},w_{50}^z+(p^\text{target,z}-w_{50}^z)*quad(\frac{i}{40})\} \nonumber \\
    &\text{where } quad(x) = 1 - (x-1)^2, \forall i \in \{0,1,...,40\} \nonumber \\
    \tau_{i+50}^\text{wrist} &= \{p^\text{target,x}, p^\text{target,y},p^\text{target,z}\} \forall i \in \{41,42...70\} \nonumber \\
    \tau_{i+50}^\text{wrist} &= \{c^x - R \sin(a_i) - h \cos(a_i),c^y + R \cos(a_i) \\
    &\ \ \ \ \ \ \ \ \ \ \ \ \ \ \ \ - h \sin(a_i),p^\text{target,z}\} \nonumber \\
    &\forall i \in \{71,72...110\} \text{ where } c^x = p^\text{target,x} + h, \\
    &\ \ \ \ \ \ \ \ \ \ \ \ \ \ \ \ c^y = p^\text{target,y}-R, a_i = \frac{i-70}{40} \nonumber \\
    \tau_{i+50}^\text{wrist} &= \{p^\text{target,x} - 0.2,p^\text{target,y},p^\text{target,z}\} \\
    &\ \ \ \ \ \ \ \ \ \ \ \ \ \ \ \ \forall i \in \{111,112,...,146\} \nonumber \\
\end{align*}

\subsection{Trajectory filtering}

As \modelname~does not consider scene elements to avoid collision with like the platform on which the target object is kept, some of the generated trajectories turn out to be infeasible to be used for reference trajectory in stage 2. Also, some of the generated trajectories empirically turned out to be very dynamically infeasible in practice to be tracked by RL in stage 2. Hence, we filter out trajectories based on some heuristics like reject a trajectory if it collides with the scene environment, reject if it bends it's waist too much etc. in the trajectory. Specifically, these are the conditions used to filter out trajectories:-
\begin{itemize}
    \item Reject if torso angle with $x$ axis is larger than $\beta^\text{torso}$ i.e. reject if $\arccos({axis_x^\text{torso,x}}) > \beta^\text{torso}$ where $axis_z^\text{torso}$ is the $x$ axis of the torso. This is to reject reference trajectories where the robot bends too much or turns around which is undesired for tasks in question
    \item Reject if pelvis height is below a certain threshold $\beta^\text{pelvis}$. This is to reject reference trajectories where robot squats too much for tasks it is not needed to.
    \item Reject if any part of the body collides with the scene
\end{itemize}

The specific thresholds, $\beta^\text{torso}$ and $\beta^\text{pelvis}$ for tasks that we do filtering and number of filtered trajectories from total sampled are given in Table \ref{tab:filter}. For all other tasks except \texttt{Open Drawer} and \texttt{Open Door}, only $1$ trajectory is manually selected for training which can be easily scaled up with careful engineering for their automated filtering. For \texttt{Open Drawer} and door tasks, as trajectories are obtained by optimization of right arm, it does not need filtering. It must be noted that these heuristic-based filtering is only required because diffusion-generated trajectories are unfit for tracking. This maybe mainly because diffusion model sees out-of-distribution samples during annealing. With more data, the need for filtering out samples maybe eliminated.   

\begin{table}[]
\centering
\caption{\# of filtered trajectories and constants used for filtering}
\label{tab:filter}
\begin{tabular}{lcccc}
\toprule
Task & \#Trajs before & \#Trajs after & $\beta_\text{torso}$ & $\beta_\text{pelvis}$ \\\midrule
\texttt{Pick}          & 100                                                                 & 67                                                                 & $\pi/4$              & 0.6                   \\ 
\texttt{Precise Punch} & 100                                                                 & 100                                                                & $\pi/4$              & 0.6                   \\ 
\texttt{Precise Kick}  & 100                                                                 & 66                                                                 & $\pi/2$              & 0.5                   \\ 
\texttt{Press Button}  & 100                                                                 & 96                                                                 & $\pi/3$              & 0.5                   \\ \bottomrule
\end{tabular}
\end{table}

\subsection{Trajectory refinement}

The trajectories obtained after re-targeting and filtering do not start from a same pose. This is a characteristic of \modelname~that it only restricts the $p^\text{ref root, x}$, $p^\text{ref root, y}$ and $\theta^\text{yaw}$ to start from origin at $t=0$ but the generated motion could start from any pose at $t=0$. However, for RL training in stage 2, we need reference trajectories to start from a fixed pose and point. Hence, in each generated reference trajectory, we prepend a trajectory that starts from a fixed default joint pose at fixed pose of the root and interpolates to the start joint pose and root pose of the generated trajectory. Specifically, we append $N^\text{init}=20$ frames at the beginning of each motion of which for the first $10$ frames the trajectory remains static and next $10$ frames to interpolate to $\alpha_0$, to obtain refined reference trajectory, $\alpha^\text{refined}$ consisting of $216$ frames. 

Further, we refine motions by disabling movement of the non-functional arm based on the task to avoid it's unnecessary movement in the reference trajectory. Specifically, we set the joint angles of the left arm and right arm group joints denoted as $G^\text{left arm}$ and $G^\text{right arm}$ as in Table \ref{tab:hand_states} to their default values along all time steps. The default joint angles. The default values are given in Table \ref{tab:joint_gains_columnwise}. The specific group that was refined to set to default joint angles for each task is given in Table \ref{tab:hand_states}. The tasks not included in table do not have this refinement applied.

\subsubsection{Special case for \texttt{Pick} task}

Last, specifically for \textit{\texttt{Pick}} task, we observed that as our robot is shorter ($1.32m$) than the SMPL model used to train generative model on ($1.74m$), most motions in the dataset are of the human picking object from platform which is located near or below it's waist while for the shorter robot, the platform is more close to it's shoulder above it's waist. Hence, most generated motion when the target point is higher for  go through the platform. Hence, specifically for the \textit{\texttt{Pick}} task, we add an additional refinement layer before passing them for filtering to minimally modify the reference trajectory to avoid collision of right hand with the platform. Specifically, we solve the following optimization problem through gradient descent with joint angle variables initialized from the reference trajectory:-

\begin{align*}
    q^\text{ref,*} = argmin &\Sigma_{t=\{\Delta t,...,(L-1)\Delta t\}} |(\lVert p_t^\text{right hand,*} - p_{t-1}^\text{right hand,*}\rVert_2 \nonumber \\
    &\ \ \ \ \ \ \ \ - \lVert p_t^\text{right hand} - p_{t-1}^\text{right hand}\rVert_2)| \nonumber \\
    &s.t.\ \  d(p_t^\text{right hand})=0 \nonumber
\end{align*}

where $d(.)$ is a function that maps 3D points to their closest distance from the free space. The rationale for choosing relative distance difference from reference as the objective is to preserve the smoothness of the motion in the generated trajectory but just modify the trajectory minimally to avoid collision of right hand with the platform. That is if the reference trajectory was to slow down while approaching the object as humans do, that slowing down will still be preserved in the trajectory after modifying it to avoid collision. If after making this refinement, if some other body part like feet in the reference trajectory makes collision with the platform, that trajectory is rejected.

\subsection{Sim2Real}

For sim2real trajectory generation and refinement, we follow a similar design to their corresponding sim tasks with some changes in prompt and trajectory refinements that we will discuss in this section. There are $5$ tasks that we put on hardware robot. Specific prompts used for them are listed in Table \ref{tab:text_prompts_real}. The generated trajectories are also slowed down by a certain factor as also reported in Table \ref{tab:text_prompts_real}. This is to ensure safety of the motion and executed and minimize the sim-real gap by making the motion slower. It was empirically observed that with the same speed, due to some sim-to-real gap, the executed trajectories exhibited less success rates for precise tasks like pick where the object got pushed a little bit and thus failed to be grasped as the position is not updated in the current design. This slowing down was not required for \texttt{Squat} and \texttt{Bimanual Pick} tasks. In addition to the standard refinements as made in the sim versions of these tasks (except \texttt{Squat}), the specific spatial control points and refinements for each task are as follows:-

\subsubsection{\texttt{Pick}}

For each reference trajectory:- We sample a target point, $p^\text{target} = \{p^x \in \mathcal{U}(0.25,0.35), p^y \in \mathcal{U}(-0.3,0.0),p^z=1.1 \}$. Target time-step is chosen to be, $t_g'=30$. Then, we set spatial control signal for wrist as follows:-
\begin{align*} \notag
    & \lambda^\text{right\_wrist}_i = p^\text{target} \forall i \in \{t_g',...,t_g'+20\} \nonumber \\
    & \lambda^\text{right\_wrist}_i = (p^\text{target,x},p^\text{target,y},p^\text{target,z}+0.2) \\
    & \ \ \ \ \ \ \ \ \ \ \ \ \ \ \forall i \in \{t_g'+20,...,t_g'+40\} \nonumber \\
\end{align*}

We also set the target spatial points for elbow to encourage generating trajectories where the object is grabbed from side as follows:-
\begin{align*}
    \lambda^\text{elbow}_i = &(p^\text{target,x},p^\text{target,y}-0.26\cos(\frac{\pi}{4}),p^\text{target,z}-0.26\sin(\frac{\pi}{4})) \nonumber\\
    &\forall i \in \{t_g',...,t_g'+20\} \nonumber\\
\end{align*}

Of the generated trajectories, we refine them to avoid collision with the platform, plus we also add an additional cost to bring the trajectories close to their corresponding goal points for which the trajectories were generated. 

\begin{align*}
    q^\text{ref,*} = argmin &\Sigma_{t=\{\Delta t,...,(L-1)\Delta t\}} |(\lVert p_t^\text{right hand,*} - p_{t-1}^\text{right hand,*}\rVert_2 \nonumber \\
    & - \lVert p_t^\text{right hand} - p_{t-1}^\text{right hand}\rVert_2)| \nonumber  \\&+ M_t \lVert p_t^\text{right wrist} - \lambda_t^\text{right wrist,'}\rVert\\
    &s.t.\ \  d(p_t^\text{right hand})=0 \nonumber
\end{align*}

Where $M_t = 1$ for $t \in \{0,t_g',t_g'+1,...,t_g'+40\} \text{ and } 0 \text{ otherwise}$. This is to mask for timesteps where the goal is specified. Idea is to minimally modify the trajectories such that they follow the corresponding goal. This refinement was required to ensure smooth policies. We also set the lower body (groups $G^\text{left leg}$ and $G^\text{right leg}$) to fixed default joint angles and adjust the root height such that the feet touches the ground. This is to enforce that the motion is stand still.

\subsubsection{\texttt{Precise Punch}}

For each reference trajectory:- We sample a target point, $p^\text{target} = \{p^x \in \mathcal{U}(0.6,0.65), p^y \in \mathcal{U}(-0.2,0.1),p^z=\mathcal{U}(1.15,1.4) \}$. Target time-step is chosen to be, $t_g'=30$. Then, we set spatial control signal for wrist as follows:-
\begin{align*} \notag
    & \lambda^\text{right\_wrist}_i = p^\text{target} \forall i \in \{t_g',...,t_g'+20\} \nonumber \\
    & \lambda^\text{right\_wrist}_i = (p^\text{target,x},p^\text{target,y},p^\text{target,z}+0.2) \\
    & \ \ \ \ \ \ \ \ \ \ \ \ \ \ \forall i \in \{t_g'+20,...,t_g'+40\} \nonumber \\
\end{align*}

Of the generated trajectories, we refine them to bring the trajectories close to their corresponding goal points for which the trajectories were generated. 

\begin{align*}
    q^\text{ref,*} = argmin &\Sigma_{t=\{\Delta t,...,(L-1)\Delta t\}} |(\lVert p_t^\text{right hand,*} - p_{t-1}^\text{right hand,*}\rVert_2 \nonumber \\
    & - \lVert p_t^\text{right hand} - p_{t-1}^\text{right hand}\rVert_2)| \nonumber  \\&+ M_t \lVert p_t^\text{right wrist} - \lambda_t^\text{right wrist,'}\rVert\\
    &s.t.\ \  d(p_t^\text{right hand})=0 \nonumber
\end{align*}

Where $M_t = 1$ for $t \in \{0,t_g'\} \text{ and } 0 \text{ otherwise}$. This is to mask for timesteps where the goal is specified. Idea is to minimally modify the trajectories such that they follow the corresponding goal. This refinement was required to ensure smooth policies. We also set the lower body (groups $G^\text{left leg}$ and $G^\text{right leg}$) to fixed default joint angles and adjust the root height such that the feet touches the ground. This is to enforce that the motion is stand still.

\subsubsection{\texttt{Bimanual Pick}}

We follow the exact same spatial control points as the sim to generate the trajectory. We additionally refine to dismiss any feet slipping by using IK to adjust $G^\text{left leg}$ and $G^\text{right foot}$ to bring left and right foot to their positions at $p_0^\text{left foot}$ and $p_0^\text{right foot}$. Other than this, we set the $roll$ and $yaw$ of root to $0$ for symmetry.

\subsubsection{\texttt{Open Drawer}}

We follow the exact same procedure as the open drawer trajectories for sim.

\subsubsection{\texttt{Squat}}

For each trajectory, we sample a trajectory with target point, $p^\text{target} = \{p^x=-0.15,p^y=0.0,p^z=\mathcal{U}(0.4,0.6)\}$. Target time-step is chosen to be $t_g'=100$. We set the spatial control signal for pelvis as: $\lambda^\text{pelvis}_i = p^\text{target} \text{ for } i=t_g'$. We additionally refine to dismiss any feet slipping by using IK to adjust $G^\text{left leg}$ and $G^\text{right foot}$ to bring left and right foot to their positions at $p_0^\text{left foot}$ and $p_0^\text{right foot}$. Other than this, we set the $roll$ and $yaw$ of root to $0$ for symmetry.

\section{RL Training}

\subsection{Model architecture}

We use a simple fully-connected MLP to represent both our policy (or actor) and critic for each task. The network architecture for the actor and critic has the following hidden layers: $(512, 256, 256)$. Also, we use the same observations for policy and critic as opposed to asymmetric actor-critic setup in \cite{Pinto2017AsymmetricAC}. 

\subsection{Environment parameters}

We randomize the object/target location (by sampling different trajectories in some tasks), object mass, friction of surface/object. All of these variations for each task are listed in Table \ref{tab:env_params}.

\begin{table}[]
\centering
\caption{Environment randomization parameters for each task}
\label{tab:env_params}
\begin{tabular}{l l l}
\toprule
Task                          & Friction             & Mass of object  \\ \midrule
\texttt{Pick}                          & $\mathcal{U}(0.7,1)$ & $\mathcal{U}(0.1,1)$           \\ 
\texttt{Precise Punch}                 & $\mathcal{U}(0.7,1)$ & -                              \\ 
\texttt{Precise Kick}                  & $\mathcal{U}(0.7,1)$ & -                              \\ 
\texttt{Press Button}                  & $\mathcal{U}(0.7,1)$ & -                              \\ 
\texttt{Jump}                          & $\mathcal{U}(0.7,1)$ & -                              \\ 
\texttt{Sit}                           & $\mathcal{U}(0.7,1)$ & -                              \\ 
\texttt{Bimanual} \texttt{Pick}                & $\mathcal{U}(0.7,1)$ & $\mathcal{U}(0.1,5)$           \\ 
\texttt{Pick} \texttt{from Ground (Side Grasp)} & $\mathcal{U}(0.7,1)$ & $\mathcal{U}(0.1,1)$           \\ 
\texttt{Pick} \texttt{from Ground (Top Grasp)}  & $\mathcal{U}(0.7,1)$ & $\mathcal{U}(0.1,0.5)$         \\ 
\texttt{Pick} \texttt{and Place}                & $\mathcal{U}(0.7,1)$ & $\mathcal{U}(0.1,0.5)$         \\ 
\texttt{Open Drawer}                   & $\mathcal{U}(0.7,1)$ & -                              \\ 
\texttt{Open Door}                     & $\mathcal{U}(0.7,1)$ & -                              \\ \bottomrule
\end{tabular}
\end{table}

\subsection{Task-specific sparse rewards}

We add task-specific sparse rewards that are chosen to be indicative of whether the task is successful as well. These rewards are listed in Table \ref{tab:task_sparse}. For each task, we have a body part link, $b$ that maybe used in the definition of the sparse rewards and the time $t_g$ when the interaction is supposed to happen.   

\subsection{Task-specific dense rewards for \emph{TaskOnly+}}

We also add a task-specific dense reward to encourage pre-grasp/pre-approach pose for the object/goal respectively. This task-specific reward is adapted from \cite{luo2024omnigrasp} and is defined as follows:-

$r_t^\text{dense} = \lVert p_{t-\Delta t}^\text{b} - p_{t_g}^\text{ref,b} \rVert_2 - \lVert p_t^\text{b} - p_{t_g}^\text{ref,b} \rVert_2$. 

The specific body parts used in each task are listed in Table \ref{tab:task_sparse}.

\begin{table}[]
\caption{Task-specific sparse rewards for each task}
\label{tab:task_sparse}
\resizebox{\textwidth}{!}{%
\begin{tabular}{l l l l}
\toprule
Task & Body part link, b & Sparse reward, $r_t^\text{sparse}$ & Description \\ \midrule
\texttt{Pick} & right\_wrist\_yaw\_link & $1_{h_t^\text{object}>h^\text{thres}}1_{t \ge t_b}$ & $h^\text{thres}=0.95$, $h_t^\text{object}$ is height of object \\ 
\texttt{Precise Punch} & right\_wrist\_yaw\_link & $1_{\lVert p_t^b - p^\text{goal} \rVert_2<d^\text{thres}} 1_{t \ge t_b-0.1}1_{t \le t_b+0.1}$ & $d^\text{thres}=0.05$ \\ 
\texttt{Precise Kick} & right\_ankle\_roll\_link & $1_{\lVert p_t^b - p^\text{goal} \rVert_2<d^\text{thres}} 1_{t \ge t_b-0.1}1_{t \le t_b+0.1}$ & $d^\text{thres}=0.1$ \\ 
\texttt{Press Button} & right\_wrist\_yaw\_link & $1_{\lVert p_t^b - p^\text{goal} \rVert_2<d^\text{thres}} 1_{t \ge t_b-0.1}1_{t \le t_b+0.1}$ & $d^\text{thres}=0.05$ \\ 
\texttt{Jump} & pelvis & $1_{\lVert p_t^b - p^\text{goal} \rVert_2<d^\text{thres}} 1_{t \ge t_b-0.1}1_{t \le t_b+0.1}$ & $d^\text{thres}=0.1$ \\ 
\texttt{Sit} & pelvis & $1_{\lVert p_t^b - p^\text{goal} \rVert_2<d^\text{thres}} 1_{t \ge t_b}$ & $d^\text{thres}=0.05$ \\ 
\texttt{Bimanual} \texttt{Pick} & right\_wrist\_yaw\_link, left\_wrist\_yaw\_link & $1_{h_t^\text{object}>h^\text{thres}}1_{t \ge t_b}$ & $h^\text{thres}=0.7$, $h_t^\text{object}$ is height of object \\ 
\texttt{Pick} \texttt{from Ground (Side Grasp)} & left\_wrist\_yaw\_link & $1_{h_t^\text{object}>h^\text{thres}}1_{t \ge t_b}$ & $h^\text{thres}=0.2$, $h_t^\text{object}$ is height of object \\ 
\texttt{Pick} \texttt{from Ground (Top Grasp)} & right\_wrist\_yaw\_link & $1_{h_t^\text{object}>h^\text{thres}}1_{t \ge t_b}$ & $h^\text{thres}=0.3$, $h_t^\text{object}$ is height of object \\ 
\texttt{Pick} \texttt{and Place} & right\_wrist\_yaw\_link & $1_{\lVert p_t^\text{object} - p^\text{goal} \rVert_2<d^\text{thres}} 1_{t \ge t_b}$ & $d^\text{thres}=0.1$, $p^\text{goal}$ is position of the goal \\ 
\texttt{Open Drawer} & right\_wrist\_yaw\_link & $1_{a_t^\text{drawer}>a^\text{thres}}1_{t \ge t_b}$ & $a_t^\text{drawer}$ is drawer open amount, $a^\text{thres}=0.05$ \\ 
\texttt{Open Door} & right\_wrist\_yaw\_link & $1_{a_t^\text{door}>a^\text{thres}}1_{t \ge t_b}$ & $a_t^\text{door}$ is door open amount, $a^\text{thres}=0.05$ \\ 
\bottomrule
\end{tabular}%
}
\end{table}

\subsection{Reward weights for each task}

The rewards weights for each task are listed in Table \ref{tab:reward_weights}.

\begin{table*}[]
\centering
\caption{Reward weights for all tasks}
\label{tab:reward_weights}
\begin{tabular}{l lllll lllll l l}
\toprule
\multirow{2}{*}{\textbf{Task}} & \multicolumn{5}{c}{\textbf{Tracking}} & \multicolumn{5}{c}{\textbf{Smoothness}} & \multirow{2}{*}{$w_{r^\text{task,sparse}}$} & \multirow{2}{*}{$w_{r^\text{task,dense}}$} \\ 
\cmidrule(lr){2-6} \cmidrule(lr){7-11}
& $w_{r_1}$ & $w_{r_2}$ & $w_{r_3}$ & $w_{r_4}$ & $w_{r_5}$ & $w_{r_6}$ & $w_{r_7}$ & $w_{r_8}$ & $w_{r_9}$ & $w_{r_{10}}$ & & \\
\midrule
\texttt{Pick} & -0.2 & -0.05 & -0.2 & -0.2 & 0.3 & -1.5$e^\text{-7}$ & -5$e^\text{-3}$ & -0.1 & -0.5 & -1 & 0.1 & 100 \\
\texttt{Precise Punch} & -0.2 & -0.05 & -0.2 & -0.2 & 0.3 & -1.5$e^\text{-7}$ & -5$e^\text{-3}$ & -0.1 & -0.5 & -1 & 1 & 100 \\
\texttt{Precise Kick} & -0.2 & -0.1 & -0.2 & -0.2 & 0.3 & -1.5$e^\text{-7}$ & -5$e^\text{-3}$ & -0.1 & -0.15 & \begin{tabular}[c]{@{}l@{}}-1 for left,\\ -0.3 for right\end{tabular} & 1 & 100 \\
\texttt{Press Button} & -0.2 & -0.05 & -0.2 & -0.2 & 0.3 & -1.5$e^\text{-7}$ & -5$e^\text{-3}$ & -0.1 & -0.5 & -1 & 1 & 100 \\
\texttt{Jump} & -0.2 & -0.1 & -0.2 & -0.2 & 0.3 & -1.5$e^\text{-7}$ & -5$e^\text{-3}$ & -0.1 & 0 & -1 & 1 & 100 \\
\texttt{Sit} & -0.2 & -0.05 & -0.2 & -0.2 & 0.3 & -1.5$e^\text{-7}$ & -5$e^\text{-3}$ & -0.1 & -0.5 & -1 & 1 & 100 \\
\texttt{Bimanual} \texttt{Pick} & -0.2 & -0.05 & -0.2 & -0.2 & 0.3 & -1.5$e^\text{-7}$ & -5$e^\text{-3}$ & -0.1 & 0 & -1 & 0.1 & 100 \\
\texttt{Pick} \texttt{from Ground (Side Grasp)} & -0.2 & -0.05 & -0.2 & -0.2 & 0.3 & -1.5$e^\text{-7}$ & -5$e^\text{-3}$ & -0.1 & -0.5 & -1 & 0.1 & 100 \\
\texttt{Pick} \texttt{from Ground (Top Grasp)} & -0.2 & -0.05 & -0.2 & -0.2 & 0.3 & -1.5$e^\text{-7}$ & -5$e^\text{-3}$ & -0.1 & -0.15 & -1 & 0.1 & 100 \\
\texttt{Pick} \texttt{and Place} & -0.2 & -0.05 & -0.2 & -0.2 & 0.3 & -1.5$e^\text{-7}$ & -5$e^\text{-3}$ & -0.1 & -0.5 & -1 & 0.1 & 100 \\
\texttt{Open Drawer} & -0.2 & -0.05 & -0.2 & -0.2 & 0.3 & -1.5$e^\text{-7}$ & -5$e^\text{-3}$ & -0.1 & 0 & -1 & 0.1 & 100 \\
\texttt{Open Door} & -0.2 & -0.05 & -0.2 & -0.2 & 0.3 & -1.5$e^\text{-7}$ & -5$e^\text{-3}$ & -0.1 & 0 & -1 & 0.1 & 100 \\
\bottomrule
\end{tabular}
\end{table*}

\end{document}